\documentclass[a4paper]{article}

\usepackage[english]{babel}
\usepackage[utf8]{inputenc}
\usepackage{graphicx}
\usepackage[colorinlistoftodos]{todonotes}
\usepackage{hyperref}
\hypersetup{
     colorlinks=true,
     urlcolor = black, 
     linkcolor=blue,
     filecolor=blue,
     citecolor = blue,  
     }

\usepackage{xspace}

\usepackage{array}
\usepackage{booktabs}
\usepackage{tikz}

\usepackage{algorithm2e}
\RestyleAlgo{ruled} 
\SetKwComment{Comment}{/* }{ */}
\usepackage{geometry}
\usepackage{tabularx}
\usepackage{amsmath,amssymb} % define this before the line numbering.
\usepackage{booktabs}
\usepackage{nth}
\usepackage{float}
\usepackage{subcaption}

\title{\vspace{-3cm}\Large \textbf{$\beta$-Multivariational Autoencoder for Entangled Representation Learning in Video Frames \xspace}}
\providecommand{\keywords}[1]{\textbf{\textit{Keywords---}} #1}

\author{{\normalfont Fatemeh Nourilenjan Nokabadi, Robert Bergevin}}

\date{
    \small LVSN-REPARTI, Université Laval, Québec, Canada \\ [2ex]%
    \href{mailto:fanon2.at.ulaval.dot.ca}{fanon2\mbox{}@\mbox{}ulaval.ca}, 
    \href{mailto:robert.bergevin.at.ulaval.dot.ca}{robert.bergevin\mbox{}@\mbox{}ulaval.ca} \\
    }

\begin{document}
\maketitle

%In our motion modelling, the current bounding box of the object is transformed into the next one by a geometrical transformation. 
%We devise a bottleneck to estimate the posterior's parameters, i.e. $\mu', \Sigma'$. 

\begin{abstract}
      It is crucial to choose actions from an appropriate distribution while learning a sequential decision-making process in which a set of actions is expected given the states and previous reward. Yet, if there are more than two latent variables and every two variables have a covariance value, learning a known prior from data becomes challenging. Because when the data are big and diverse, many posterior estimate methods experience posterior collapse. In this paper, we propose the $\beta$-Multivariational Autoencoder ($\beta$MVAE) to learn a Multivariate Gaussian prior from video frames for use as part of a single object-tracking in form of a decision-making process. We present a novel formulation for object motion in videos with a set of dependent parameters to address a single object-tracking task. The true values of the motion parameters are obtained through data analysis on the training set. The parameters population is then assumed to have a Multivariate Gaussian distribution. The $\beta$MVAE is developed to learn this entangled prior $p = N(\mu, \Sigma)$ directly from frame patches where the output is the object masks of the frame patches. We devise a bottleneck to estimate the posterior's parameters, i.e. $\mu', \Sigma'$. Via a new reparameterization trick, we learn the likelihood $p(\hat{x}|z)$ as the object mask of the input. Furthermore, we alter the neural network of $\beta$MVAE with the U-Net architecture and name the new network $\beta$Multivariational U-Net ($\beta$MVUnet). Our networks are trained from scratch via over 85k video frames for 24 ($\beta$MVUnet) and 78 ($\beta$MVAE) million steps. We show that $\beta$MVUnet enhances both posterior estimation and segmentation functioning over the test set. Our code and the trained networks are publicly released.
\end{abstract}

\section{Introduction}

Building a successful pipeline for a machine learning method is highly dependent on the data representation. Consequently, designing a precise feature extraction technique has an impact on the method's performance. As a popular trend, training a distributed, invariant, and disentanglement representation has led researchers to develop AI models involving a set of independent variables~\cite{bengio_representation_2013}. Representation learning is aimed to facilitate the feature extraction~\cite{bengio_representation_2013}, while providing such a distributed independent setting for real-world problems is not always the case~\cite{raposo_discovering_2017}. For instance, learning to predict the continuous control parameters in dynamic models has been built on a Multivariate Gaussian Distribution (MGD) where the covariance matrix is diagonal~\cite{janner_when_2019}. However, authors in~\cite{zhang_autoregressive_2022} challenged the independence assumption of these models by using an autoregressive dynamic model for policy evaluation and optimization. Deep neural networks are capable of predicting posteriors in three forms including Variational Auto-Encoders(VAEs)~\cite{kingma_auto-encoding_2014},~\cite{rezende_stochastic_2014}, deep Auto-Regressive(AR) methods~\cite{gregor_deep_2014}, and deep generative networks with Normalizing Flows(NF)~\cite{kingma_improved_2016},~\cite{tomczak_improving_2017}. These likelihood-based generative models are aimed to use an encoder to learn a posterior distribution, when a decoder is expected to re-generate input data. In AR models and methods with NFs, a set of inevitable transformations is applied to map a simple distribution into a complex one. The sequential inherent of these methods slow down the learning process. Also, they overparameterize in the case of the complex density estimation~\cite{kingma_improved_2016}. Ultimately, in large dataset learning, these methods easily confront the posterior collapse~\cite{lucas_understanding_2019}, while the VAEs as well-known generative models are more flexible in applications like biomedical image analysis \cite{ilse_diva_2020}, image super-resolution \cite{liu_variational_2021} and abnormality detection \cite{yan_abnormal_2020}.  

%The underlying variables in the problems can be related to each other and because of those relations, learning their correlations is essential in order to address the corresponding tasks. [***Connection Sentence***]

\begin{figure*}[t]
\vskip -0.2in
\begin{center}
\centerline{\includegraphics[width=1\columnwidth]{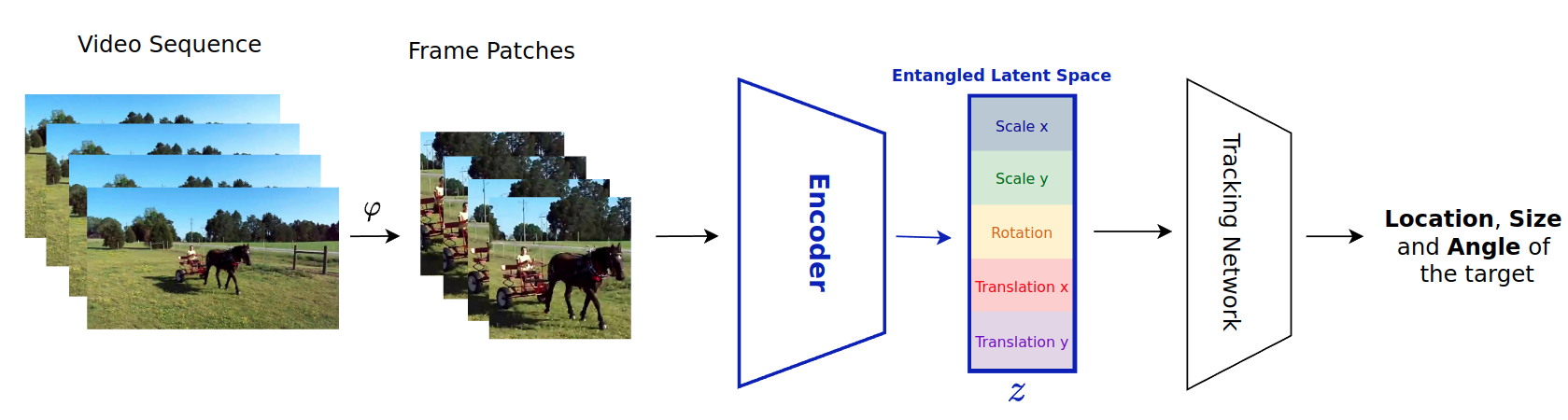}}
\caption{The general idea of our tracking method contains an encoder to place the variables into an appropriate multivariate Gaussian space where $\varphi$ is the preprocessor to crop the video frames appropriately. This paper focuses on developing and training the encoder with the bottleneck
. Once we have learned the action space, i.e. latent space, we use these multivariational inferences to form the MDP states and estimate the tracking parameters based on them.}
\label{sys}
\end{center} 
\vskip -0.2in
\end{figure*}

Following an investigation to develop an object tracker via Reinforcement Learning techniques(RL)~\cite{sutton_reinforcement_2018}, we reach a gap between the RL algorithms and our object tracking modelling. In RL-based trackers~\cite{dunnhofer_visual_2019} \cite{yun_action-decision_2017} \cite{choi_real-time_2018}, the tracking task is primarily formulated by a Markov Decision Process (MDP) in which states are video frames usually used to predict the MDP's actions, i.e. tracking parameters. However, estimating the actions directly from the video frames is not a sensible idea. Because the tracking parameters always have a different distribution from the video patches. Therefore, we proposed $\beta$MVAE to learn the tracking parameters distribution from the video patches. As indicated in Figure \ref{sys}, we will develop our tracking method such that the continuous parameters are sampled from a posterior with a non-diagonal covariance matrix estimated by the raw pixels. The two essential steps of this work contain learning to estimate the posterior from pixels and the object mask prediction by the decoder. 

In our method, we first formulate the object's motion across video frames by a geometrical transformation to calculate the next object bounding box from the current one. Then, we propose the $\beta$-Multi-Varaitional AutoEncoder ($\beta$MVAE) to learn the representation of the moving object segmentation where the bottleneck obeys an explicit prior, i.e. an MGD with a full covariance matrix. Our desired latent space has five variables $V = \{ v_1, v_2, v_3, v_4, v_5 \}$, and these correlated parameters should be chosen to address the object tracking task. This entangled space contains two scale parameters, one rotation parameter, and two translation parameters relating to the object's motion across frames, respectively. This prior $p = N(\mu, \Sigma)$ is obtained after analysing the training set's data using our proposed tracking modelling. Following that, we assume an MGD over the true values of the parameters, where $\mu$ is the 5-element mean vector corresponding to the latent variables $V$ and $\Sigma$ is its $5 \times 5$ covariance matrix. Given this prior, we created $\beta$MVAE to predict the posterior from video frames by the encoder while the segmentation map is estimated by the decoder. Our contributions are summarised as follows:

\begin{enumerate}
    \item We formulate a novel dynamic to model the single object's motion across video frames.
    \item The $\beta$Multi-Variational Auto-Encoder ($\beta$MVAE) is developed to learn a multivariate Gaussian distribution with full covariance matrix from raw pixels in addition to the object mask of the frame patches.
    %\item Our proposed methodology suggests a novel approach to reconstruct the data distribution of a multivariate Gaussian distribution. 
    \item A novel trick is introduced for the bottleneck reparameterization to map a set of the prior samples to the posterior's parameters to add the randomness in the proposed structure. 
    \item The bottleneck is directly trained by computing Kullback–Leibler (KL) divergence between the prior and the estimated posterior instead of learning the expectation of the lower bound. 
    \item We change the neural network architecture of auto-encoder to a U-Net in our proposed method. The outcomes of posterior estimation and segmentation mask creation is enhanced by the new architecture. 
\end{enumerate}

In the following sections, we review the background of posterior estimation methods, and motion modelling in videos in Section \ref{relatedworks}.  Next, we present our problem statement containing the motion modelling, bottleneck structure, reparameterization trick and the training procedure in Section \ref{pre}. The proposed method is described in Section \ref{MVAE}. Finally, the experimental results demonstrate the efficiency of our method in Section \ref{Results}.

%%________________________________________________________

\section{Related Works}
\label{relatedworks}

%In the literature, the posterior estimation is learned by either Variational Auto-Encoder (VAE) methods, or Normalizing Flows (NF) approaches. For video modeling, the object motion is modelled    **

%We briefly discussed the methods in the following. 
\subsection{Variational autoencoders}

%\textbf{VAEs}. 
The VAE models~\cite{joo_dirichlet_2020},~\cite{nazabal_handling_2020} strive to predict the conditional densities of the latent variable(s). The essential notion behind these models is to assume a target prior, and then, to estimate the closest approximation to the target likelihood. Considering independent latent variables, the disentanglement representation methods are developed to predict mean-field family of densities where the target distribution is explainable with fully factorized densities~\cite{blei_variational_2017}. However, hierarchies between variables are also considered in some cases such as ~\cite{ranganath_hierarchical_2016},~\cite{vahdat_dvae_2018}, and~\cite{vahdat_undirected_2020} where the latent variables are discrete. Discrete Variational Auto-Encoders (DVAE++)~\cite{vahdat_dvae_2018} is also proposed for a mixture of two overlapping distributions with a new variational bound to train Boltzmann machine priors. DVAE++ is developed to learn the priors for discrete variables with a continuous relaxation hierarchy. Another attempt for posterior approximation is Undirected Graphical Models (UGM)~\cite{vahdat_undirected_2020} to learn the undirected approximate posteriors with Markov Chain Monte Carlo updates. With UGM, a latent space of discrete variables is trained with the restricted Boltzman machine parameters. As an example, the UGM is also trained on a multivariate Gaussian distribution where the mean vector is $\mu = [1, 1]$ and the covariance matrix is $\Sigma = [1.1, 0.9; 0.9, 1.1]$ for two discrete variables. Although the UGM is tested using a multivariate distribution, its approach is completely different from $\beta$MVAE. It has been examined only on two discrete variables where the covariance of the data is positive. Hierarchical Variational Models (HVM) \cite{ranganath_hierarchical_2016} expanded the idea of VAE for mean-field posteriors to hierarchical posteriors. In mean-field families, the latent variables are independent from each other, while in hierarchical models, the target prior presents an arbitrary structure for the dependencies of the latent variables. The HVM contains two components including the variational likelihood representing the dependencies and the prior as the target posterior. In contrast, our method is trained for learning an MGD where the mean vector and covariance matrix are obtained after data analysis for our vision task. The $\beta$MVAE works for continuous latent variables with the arbitrary covariances. Furthermore, we succeeded in achieving the results with calculating the KL divergence instead of the expectation of the lower bound for network optimization.

Locatello et al. \cite{locatello_challenging_2019} discussed that learning meaningful features via VAEs requires supervision to some extent. Following this claim, the DIVA method~\cite{ilse_diva_2020} employed supervision in terms of domain and class labels. In $\beta$MVAE, we use supervision including class labels, segmentation mask and the prior during the training.

%For instance Discrete Variational Auto-Encoders (DVAE++) is developed to learn the priors for discrete variables with a continuous relaxation hierarchy. Authors in~\cite{vahdat_undirected_2020} tested their method using a multivariate distribution. However, their approach is completely different from $\beta$MVAE. It has been examined only on two discrete variables where the covariance of the data is positive. 

%The VAEs are widely used for the extraction of input features such as motion and appearance~\cite{yan_abnormal_2020},~\cite{fan_video_2020} for the detection of the abnormality in videos, for motion segmentation \cite{nagano_high-dimensional_2019}, and 3D shape reconstruction~\cite{nash_shape_2017}. 

%To join this trend, $\beta$MVAE is developed to extract binary object masks from video frames. 

%the original idea of VAEs without any specific sampling or gradient computation other than common deep network optimization.
%
%

%\textbf{NFs for posterior estimation}. 

\subsection{Normalizing flows}

Normalizing Flows(NFs)~\cite{dinh_density_2017},~\cite{rezende_variational_2015} are proposed to convert a sample of a simple distribution to a sample of a complex distribution over a set of invertible transformations. Because in deep learning algorithms, the invertiblity is essential to compute the back propagation. The Planar, and Radial flows~\cite{rezende_variational_2015} are introduced for posterior estimation and evaluated in a group of 2D densities. Similar to other methods, these approaches are trained and tested on image dataset such as MNIST and CIFAR-10 which are easier data to deal with than video datasets. Using bijective transformation, the real-valued non-volume preserving(RealNVP)~\cite{dinh_density_2017} is proposed to predict the posterior. The RealNVP is tested over images datasets with resolutions $32 \times 32$ and $64 \times 64$ which is simpler cases in compared to our dataset, i.e. video sequences with frame patches $256 \times 256$.  Moreover, all of these methods mainly focused on one to two variables with fully factorized distributions. The inverse Autoregeressive Flow(IAF)~\cite{kingma_improving_2017} is suggested to predict the variational Inference for high dimensional latent space by considering even variable dependencies via an autoregressive approach. In experiments, the proposed scenario of IAF is evaluated in a simplified case of a MGD using simple datasets such as MNIST and CIFAR-10.

%----------

%\textbf{Motion Modelling in Videos}. 

\subsection{Motion modelling in videos}

Several scenarios are proposed in the literature to model single moving object in video frames. Some methods follow Kalman filters theory~\cite{ljouad_hybrid_2014} developed based on physical laws related to the target's motion in order to predict the location of the moving target in the next frame in a video. Some other methods employed optical flow~\cite{yang_unsupervised_2019},~\cite{koh_primary_2017},~\cite{cheng_segflow_2017},~\cite{wang_saliency-aware_2015} to find the moving objects. The optical flow methods are established on the assumption of the brightness constancy constraint which states that the brightness of key points in two adjacent frames is the same. Other methods model the object's motion with sequential decision making~\cite{yun_action-decision_2017} which describes the object's motion in two frames with discrete parameters~\cite{yun_action-decision_2017},~\cite{huang_learning_2017}, \cite{zhong_hierarchical_2019}, or continuous parameters~\cite{chen_real-time_2018}, \cite{dunnhofer_visual_2019}, and in some cases both types~\cite{ren_deep_2018},~\cite{liu_revisiting_2018}.  The continuous parameters provide smoother steps for the object's motion modelling and because of this fact, our proposed method is developed to train an encoder for a sequential decision-making method with continuous parameters describing the object's motion in video frames.

\subsection{Saliency detection}
%\textbf{Saliency Detection}. 

Segmenting the foreground object from the background is extensively studied by classical approaches~\cite{nouri_salient_2015} to neural network based models~\cite{li_deepsaliency_2016},~\cite{chen_ef-net_2021}, \cite{liu_simple_2019}. Saliency detection is also proposed for videos~\cite{fang_devsnet_2020} by taking temporal information into account rather than visual image cues. Although our method does not rely on temporal information in forming the segmentation maps, the saliency maps are generated for each input by processing spatial features.

%----------------------------------------------------------------

\section{Preliminaries}
\label{pre}

Given input data $x$, conventional $\beta$VAE~\cite{higgins_beta-vae_2016} is aimed to train an encoder $q_{\phi}(z|x)$ for the variational inference(s) $z$ approximation which lies in the latent space and a decoder $p_{\theta}({x}|z) $ to estimate the likelihood of input data ${x}$ from $z$. To encourage the network to learn a better representation, the coefficient $\beta$ multiplies the $D_{KL}$ term in the loss function as follows:

\begin{equation}
\label{Bvae}
    \mathcal{L}_{\beta \text{VAE}} = - {\mathbb{E}}_{q_{\phi}(z|x)} \big [ \text{log} p_{\theta}({x}|z) \big ]
    + \beta D_{KL} \big( q_{\phi}(z|x) || p(z) \big)
\end{equation}

Since our network is expected to learn the segmentation mask $\hat{x}$, the likelihood is rewritten as $p_{\theta}(\hat{x}|z)$. Moreover, the reconstruction loss is named construction loss in our setting. It compares the estimated segmentation mask $\hat{x}$ with the true annotation $gt$ of the frame patch. The loss function of $\beta$MVAE is calculated as follows:

\begin{equation}
      \mathcal{L}_{\beta \text{MVAE}} = {\mathcal{L}_{cons} (\hat{x}, gt)  + \beta \mathcal{L}_{KL} }%\big)
      \label{eq1}
\end{equation}

\noindent where $\mathcal{L}_{KL} = D_{KL} \big( q_{\phi}(z|x) || p(z) \big)$ is directly computed by creating two MGDs for the estimated posterior and the prior. Also, the $\mathcal{L}_{cons}$ is defined as:

\begin{equation}
\label{Lrecons}
    \mathcal{L}_{cons}(\hat{x}, gt) = \mathcal{L}_{ce}(\hat{x}, gt) + \mathcal{L}_{\mathcal{J}}(\hat{x}, gt) 
\end{equation}

\noindent where ${x'}$ is the decoder's output, $gt$ is the annotation, $\mathcal{L}_{ce}(\hat{x}, gt)$ is the cross entropy loss and $\mathcal{L}_{\mathcal{J}}({x'}, gt) = 1 - \mathcal{J}({x'}, gt)$ is the Jaccard index loss. For cross entropy, we first calculate the weight of each class in the frame patch and then, compute the multi-class cross entropy. Also, the Jaccard index loss is added to boost the segmentation ability of the network.

% The latent space of $\beta$MVAE is an $p = N(\mu, \Sigma)$ with five variables where each one relates to one motion parameter in our motion model. Following our suggested scenario for single object motion in video frames, described in the next Section, the true values of the parameters is computed. Then, an MGD is assigned to the data population and employed as a prior for our training purpose. 

%------------------------------------------------------------------------- 

\section{Proposed Method}
\label{MVAE}

\begin{figure*}[!t]
\centering
%\vskip -0.2in
\includegraphics[width=0.8\textwidth]{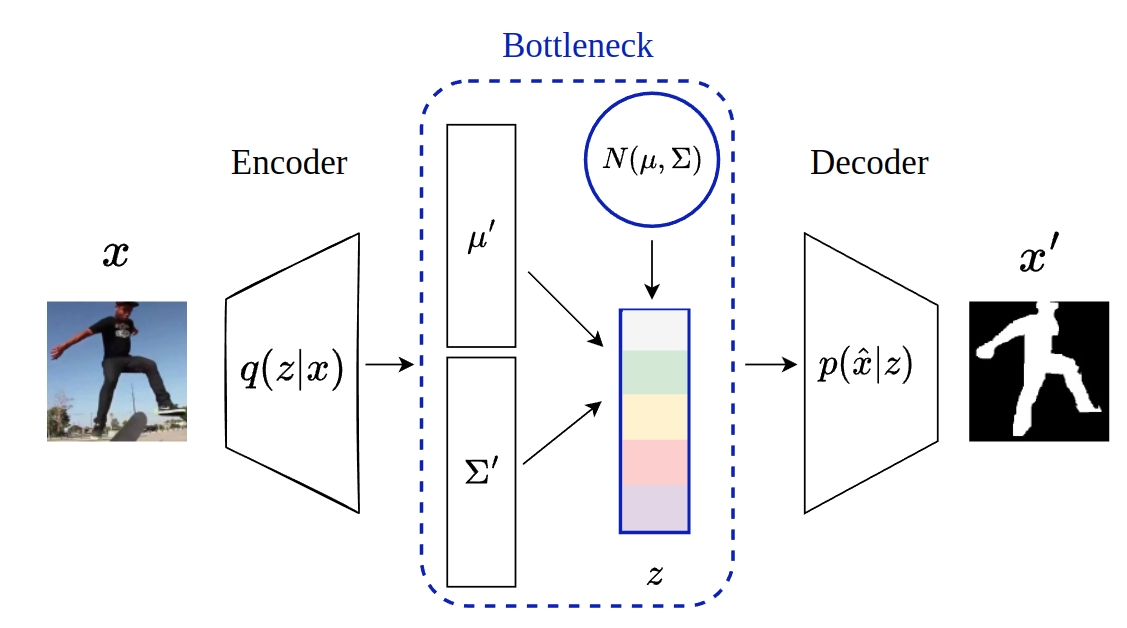}
\caption{Overview of our proposed method, in which video frames are encoded into a sample set of the posterior, while the decoder is expected to estimate the object binary mask using the multivariational inferences.}
\label{VAE_MV}
%\vskip -0.2in
\end{figure*}

We propose a novel formulation to model the object's motion across frames in videos. According to our formulation, the single moving object is transformed geometrically at each time step in the frame plane. By performing a data analysis, we obtained true values for each parameter in our motion model over the training set. Then, we assigned a MGD to our data population of the motion parameters. The $\beta$MVAE is developed to learn this MGD as the latent space, while the decoder is able to construct the segmentation mask, see Figure \ref{VAE_MV}. 

%****
%One important difference between our method and the conventional VAEs is the structure of the bottleneck. In existing VAEs, \todo{I should talk about how the loss of KL is calculated here and in other methods!}

%the bottleneck is generally designed to reproduce one variational inference per input, while in MAPD, the bottleneck infers a sample set of our target space per input. This difference provides an opportunity to rebuild a MGD per input batch and learn to follow our prior. 

In this section, we first describe our motion model for the single moving object in videos. Next, we explain the bottleneck structure and posterior's estimation per input batch. Then, we introduce a new reparameterization trick to strengthen the encoder so as to predict the posterior. Subsequently, we explain the training procedure details and finally, the $\beta$MVUnet is proposed to assist our method in learning the segmentation masks.

\subsection{Single moving object model in video sequence} 

%Unique Transformation of \textit{rbbox} across Frames}
%in one plane

\label{idea}

\begin{figure}[!t]
\centering
\includegraphics[width=0.8\textwidth]{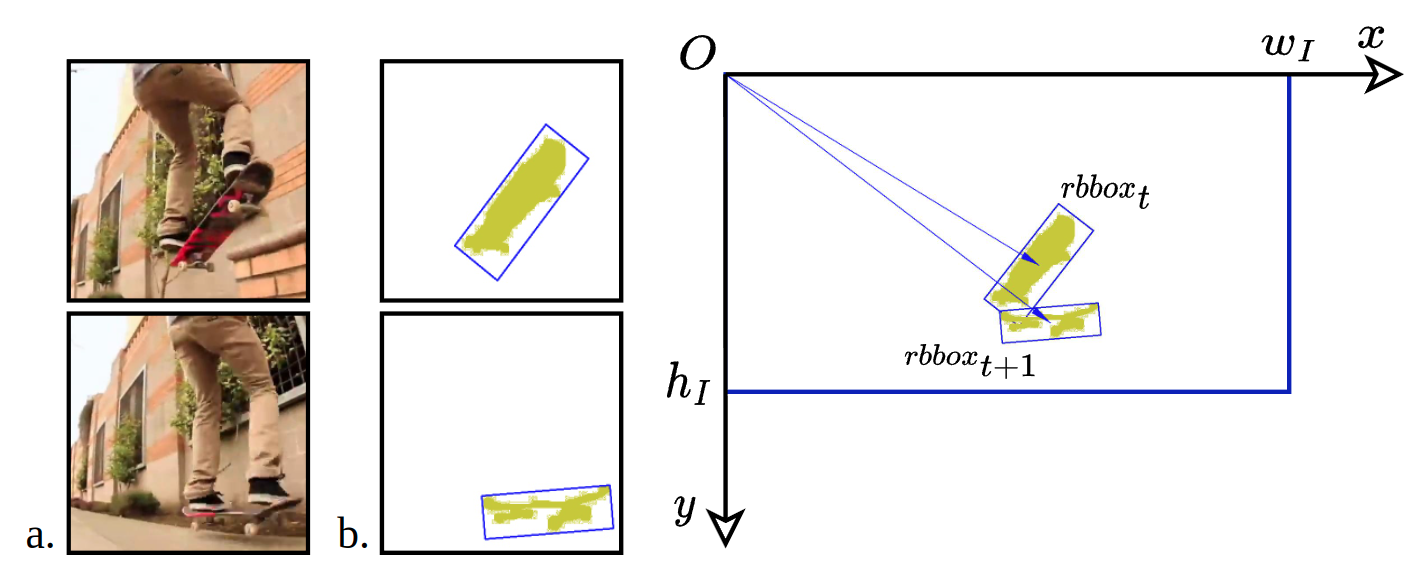}
\caption{Let us consider two successive frames \textit{rbbox}s. One can see that the \textit{rbbox} at time step $t+1$ is the transformed \textit{rbbox} at time step $t$, (a) (top) frame patch at time step $t$, and (bottom) frame patch at time step $t+1$, (b) (top) binary mask with \textit{rbbox} at time step $t$, (bottom) binary mask with \textit{rbbox} at time step $t+1$. On right, two \textit{rbbox}s are shown in the frame as the foregrounds.}
\label{assumption}
\end{figure}

In our method, the mask is used to compute the rotated bounding box (\textit{rbbox}) in each time step. Then, we employ the (\textit{rbbox})s to describe the moving object characteristics in the frame  as $\textit{rbbox} = ([x, y], [w, h], \alpha)$ representing the center position, size and angle of the \textit{rbbox}.

The underlying assumption of our method is that any motion of \textit{rbbox}s between two successive frames is decipherable by three geometric transformations: translation, counterclockwise rotation, and scale. Figure \ref{assumption} illustrates this idea for two consecutive frames. The object's position in the next frame $[x', y']$ can be calculated by transforming the current position $[x, y]$ as:

\begin{equation} \label{eqCenter}
    \begin{bmatrix}
    x' \\
    y' \\
    1
    \end{bmatrix} 
    = G_t \times \begin{bmatrix}
    x \\
    y \\
    1
    \end{bmatrix}
\end{equation}

\noindent where { $G_t$} is:

\begin{equation}
\label{gtMat}
G_t = \begin{bmatrix}
\Delta s_x \times {cos \theta} & -\Delta s_y \times {sin \theta} & \Delta x\\
\Delta s_x \times sin \theta & \Delta s_y \times cos \theta & \Delta y\\
0&0&1\\
\end{bmatrix}
\end{equation}

The key idea of our tracking system is predicting $G_t$ at time step $t$ by processing the video frames. According to Equation \ref{gtMat}, we need to predict five motion parameters $[\Delta s_x, \Delta s_y, \theta, \Delta x, \Delta y]$ to predict the $G_t$ at each time step $t$. The translation parameters are normalized regarding the frame dimensions as $\Delta x = t_x / w_I$ and $\Delta y = t_y / h_I$ where $w_I$, $h_I$, $t_x$ and $t_y$ are respectively the width of the frame, height of the frame, and raw translation values in $x$ axis, and $y$ axis. Furthermore, for every two successive frames, there is one unique $G_t$ to transform the \textit{rbbox} from the current step to the next one. The proof of $G_t$ uniqueness is available in the Appendix.

%To form the covariance matrix of the input data, we have to calculate the relations between variables each time during training. We developed $\beta$MVAE such that each input data is mapped into a sample set of the latent space instead of one variational inference. In this way, the bottleneck represents a sample set $z$ of the latent space. Using this sample set $z$, the decoder is able to predict the object binary mask.

%------------------------------------------------------------------------- 

\subsection{Bottleneck structure and distribution}
\label{ddKL}

\begin{figure*}[!t]
%\vskip -0.2in
\centering
\includegraphics[width=\textwidth]{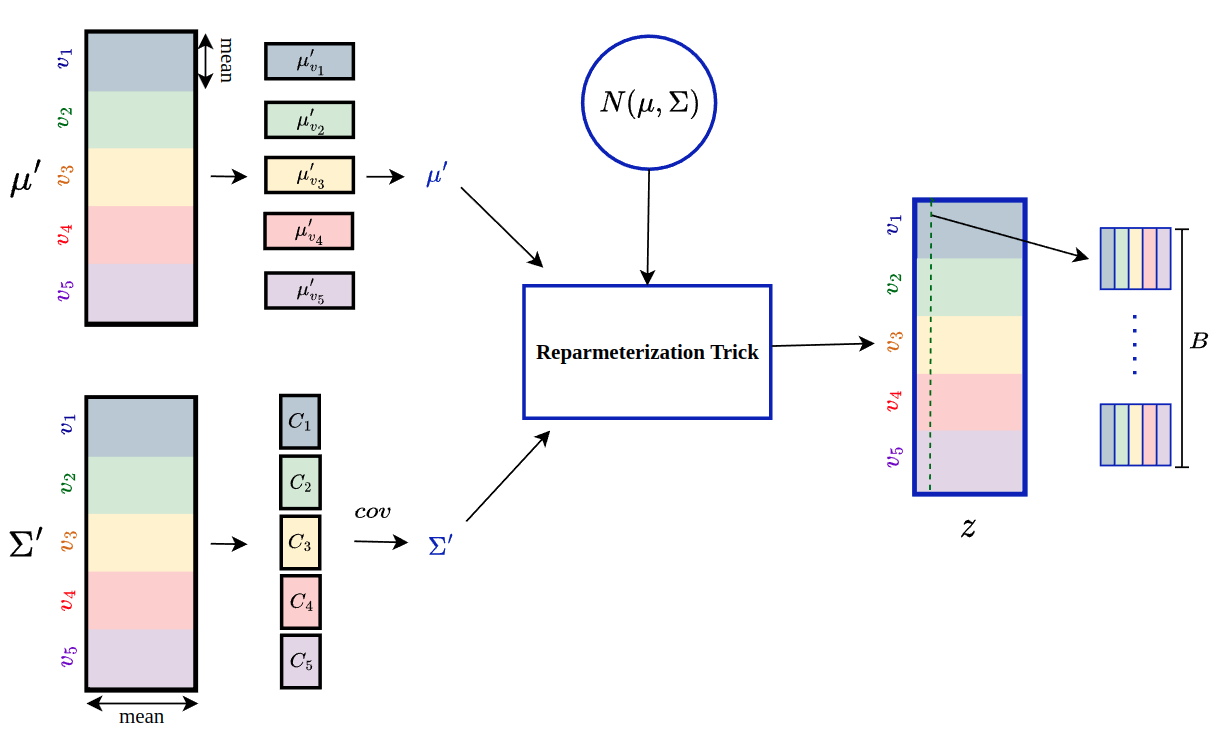}
\caption{The parameters of the posterior, $\mu'$ and $\Sigma'$, computed by two fully-connected layers following the encoder's output. The mean vector and co-variance matrix are calculated by assigning every $50$ neurons to one variable. The reparameterization trick adds random samples of prior to the bottleneck. If the bottleneck $z$ is cropped in the axis of the batch size $B$ (the dashed line), a sample set is created containing $B$ groups of variational inferences, each one containing $50$ rows of $5$-element samples representing the encoder's posterior.}
\label{dd}
%\vskip -0.2in
\end{figure*}

For each batch of the frame patches, the $\beta$MVAE's encoder infers two sets of outputs, one to generate the mean vector $\mu'$ and the other for the covariance matrix $\Sigma'$ estimation. These outputs are derived from two fully connected layers following the encoder's output. In the bottleneck, we aim to learn a sample of data representing the covariance matrix and the mean vector, depicted in Figure \ref{dd}. 
The fully connected layer $\mu'$ is divided into five parts of 10 neurons, as it is illustrated in Figure \ref{dd}. For a data batch of the size of $(B \times 256 \times 256 \times 3)$, the mean vector is constructed by obtaining the average of five parts of $\mu'$. Unlike the mean vector, the covariance matrix is not estimated element by element due to the difficulty of holding a covariance matrix properties, i.e. forming a symmetric and semi-definitive matrix. Therefore, a data population is employed to reproduce the covariance matrix of the input data. The fully-connected layer $\Sigma'$ is partitioned into five parts of $10$ neurons, as in the first step of the mean vector generation. Next, the norm of each part is computed in the axis of the batch, which results in five vectors of length $1 \times 10$. Subsequently, we estimated the covariance matrix of these five samples. In this way, the consequent covariance matrix is a symmetric and semi-definitive matrix. As illustrated in Figure \ref{dd}, the bottleneck $z$ contains $B$ sets of variational inferences of size $5 \times 10$ where $B$ is the batch size, $5$ represents the number of variables in the prior, and $10$ is the number of samples to create the posterior which is close to the prior in terms of KL divergence.

\subsection{Reparameterization trick}

\begin{table}[t]
%\small
\caption{Lower Bound (LB) and Upper Bound (UB) of the latent distributions. These numbers are rounded floating points. }
\label{ULB}
\begin{center}
    \begin{tabularx}{0.8\textwidth} { 
       >{\centering\arraybackslash}X 
       >{\centering\arraybackslash}X 
       >{\centering\arraybackslash}X
       >{\centering\arraybackslash}X  }
     \hline 
     variables & $\mathcal{N}(\mu, \sigma)$ & LB$(\mu - \sigma)$ & UB$(\mu + \sigma)$ \\ [5pt]
     \hline
    $v_1$  & {$\mathcal{N}(1.06, 0.52)$} & $0.55$ & $1.59$ \\ [5pt]
    $v_2$ &  {$\mathcal{N}(1.06, 0.54)$} & $0.53$ & $1.61$ \\ [5pt]
    $v_3$ & $\mathcal{N}(0, 0.43)$ & $-0.43$ & $0.43$ \\ [5pt]
    $v_4$ & $\mathcal{N}(0.07, 0.34)$ & $-0.27$ & $0.41$ \\ [5pt]
    $v_5$ & $\mathcal{N}(0.08, 0.74)$  & $-0.66$ & $0.82$\\ [5pt]
    \hline
    \end{tabularx}
\end{center}
%\vskip -0.2in
\end{table}

%explained in Subsection \ref{ddKL}

In traditional VAEs, the decoder is expected to sample from the true posterior $z \sim p(z|x)$. The reparameterization trick is applied to maintain the VAEs stochastically. In other words, it adds randomness in the VAEs by first sampling from a noise distribution $\epsilon \sim N(0, 1)$. After that, the VAEs reparameterize this sample using the estimated posterior $q(z|x) = N(\mu, \sigma)$ by $z = \mu + \sigma \varepsilon $. In our proposed method, we suggest a similar technique to reparameterize the random prior samples using the estimated posterior $N(\mu', \Sigma')$. In this way, the randomness comes from the prior that the bottleneck is expected to learn. Then, we linearly transform the random samples into the data distribution. To ease this process, we assumed that the prior contains five independent normal distributions, as shown in Table \ref{ULB}, which indicates the priors of the latent variables.  The covariance between variables are left for the networks to learn from the KL divergence term in the loss function. For each distribution, the upper and lower bounds are computed as $\mu + \sigma$ and $\mu - \sigma$. As described in Subsection \ref{ddKL}, the parameters of the posterior is calculated as the mean vector $\mu'$ and the covariance matrix $\Sigma'$. Using the diagonal elements of $\Sigma'$, the variable's variances $\sigma'_i$s are computed in this step. Then, we obtain the coefficients of the linear transformation to map the prior samples $p$ to the encoder's posterior $q$ using the lower and upper bounds of each distribution. The coefficients are calculated as $a_i = \frac{\sigma'_i}{\sigma_i} $ and $b_i = \mu'_i - \mu_i \times \frac{\sigma'_i}{\sigma_i}$ for the $i^{th}$ part of the bottleneck, therefore:

\begin{equation} \label{finalz}
    z_i = {a_i} \varepsilon_i + b_i
\end{equation}

\noindent where $z_i$ is forming the i-$th$ part of our latent set of samples $z$. Given a sample from the prior $\varepsilon \sim N(\mu, \Sigma)$ of size $B \times 50$, we first divide it into five parts of $10$ neurons. Then, for part $i$, the transformation in Equation \ref{finalz} is applied to force the random samples to have the encoder's posterior characteristics. 

\subsection{Training procedure}
 \label{train}

Drawing a random batch $\mathcal{X}_{batch}$ from the training set $\mathcal{X}_{train}$, the training procedure of the $\beta$MVAE, Algorithm \ref{alg:1}, is started by feeding $\mathcal{X}_{batch}$ to the encoder $q(z|x)$. Next, the mean and covariance of the posterior are calculated as $\mu'$ and $\Sigma'$ by two fully connected layers. The posterior of $\beta$MVAE is estimated as $q=N(\mu', \Sigma')$. Here, the KL divergence loss is computed as $\mathcal{L}_{kl} = D_{\mathcal{KL}} (p||q)$ between the prior and the predicted posterior. Next, the multivariational inferences $z$ are created using our new reparametrization trick and fed to the decoder to predict the segmentation map of the frame patch. The network's parameters are updated after one feed-forward pass using the loss function. The backbone of $\beta$MVAE for the encoder and the decoder is ResNet18 \cite{he_deep_2016}, while the mean and covariance values are computed by two fully connected layers. We employ Leaky ReLU as the activation function in all layers of $\beta$MVAE's encoder and decoder. The training set is prepared and preprocessed using the YouTube-VOS dataset \cite{xu_youtube-vos_2018} as explained in detail in Appendix.

\begin{algorithm}[!t]
%\small
\caption{Training $\beta$MVAE for estimating the posterior and binary object mask} \label{alg:1}
\DontPrintSemicolon
\KwData{ \; 
        \hskip1em $q(z|x)$: encoder network\;
        \hskip1em $p(\hat{x}|z)$: decoder network \;
        \hskip1em $\mathcal{X}_{train}$: training dataset \;
        \hskip1em $FC1$: mean layer $\mu'$ \;
        \hskip1em $FC2$: covariance layer $\Sigma'$ \;
        \hskip1em $p = \mathcal{N}(\mu, \Sigma)$: prior}

\KwResult{ \;
        \hskip1em $z$: multivariational inferences \; 
        \hskip1em ${\hat{\mathcal{X}}_{batch}}$: segmentation mask } %\;
\While{not converged}{
    $\mathcal{X}_{batch} \gets$ sample a batch from $\mathcal{X}_{train}$ \;
    $\varepsilon \gets$ sample from the prior $\mathcal{N}(\mu, \Sigma)$ \; %\;
    $\mu' \gets \text{FC1} \big ( q(z|\mathcal{X}_{batch}) \big)$ \;
    $\Sigma' \gets \text{FC2} \big ( q(z|\mathcal{X}_{batch})  \big)$ \;
    $\sigma'_i \gets$ $i^{th}$ element of the diagonal of $\Sigma'$ \;
    $\sigma_i \gets $ from Table \ref{ULB} \;
    \For{$i=1, ..., 5$}{
        $a_i \gets \frac{\sigma'_i}{\sigma_i} $ \;
        $b_i \gets \mu'_i - \mu_i \times \frac{\sigma'_i}{\sigma_i}$ \;
        $\varepsilon_i \gets $ select $i^{th}$ part of $\varepsilon$ \;
        $z_i \gets {a_i} \varepsilon_i + b_i $, Eq. \ref{finalz} \;
        }
    $q \gets $ compute the posterior $\mathcal{N}(\mu', \Sigma')$ \;
    $\mathcal{L}_{kl} \gets D_{\mathcal{KL}} (p || q) $ \;
    ${\hat{\mathcal{X}}_{batch}} \gets p({\hat{\mathcal{X}}_{batch}}|z)$ \;
    evaluate $\mathcal{L}_{cons}$, Eq. \ref{Lrecons} \;
    Update networks by $\mathcal{L}_{\beta \text{MVAE}}$ (Eq. \ref{eq1}) \;
}
\end{algorithm}

%backpropagation of the total loss 

Our dataset consists of 3471 video sequences split into $80\%$ and $20\%$ as the training and validation sets. In total, we trained our network with 68131 frames as the training set and 17028 frames as the validation set. The learning is performed with a single GPU RTX 3090 for $920$ epochs in $24$ days, while the validation is performed every $20$ epochs. The batch size is fixed at $32$. The optimization was conducted via RAdam~\cite{liu_variance_2021} scheduled according to the cyclical learning rate policy~\cite{smith_cyclical_2017} between $5e^{-5}$ and $1e^{-4}$. The coefficient $\beta$ in Equation \ref{eq1} is set to $\beta = 5$ similar to the $\beta$VAE~\cite{higgins_beta-vae_2016} experiment for 3D chairs. Choosing $\beta > 1$ helps the network to learn the posterior more efficiently~\cite{higgins_beta-vae_2016}. During the training, the KL divergence term fluctuated, especially when the construction loss $\mathcal{L}_{cons}$ began to decrease. Further details concerning the training phase are discussed in the appendix.
%However, the network began to learn the segmentation task once the $\mathcal{L}_{KL}$ was fixed.

\subsection{$\beta$Multi-Variational U-Net}
\label{ablation}
%$\beta$Multivariational UNet

\begin{table}[t]
\caption{$\beta$MVAE and $\beta$MVUnet network characteristics. The \textbf{outc} is the output layer that maps encoder results into $66$ classes. The \textbf{ResNet18$^\dagger$} is the ResNet18 architecture plus one extra convolutional layer. \textbf{Linear} refers to the fully connected layer without activation function which means linear mapping.}
\label{Arch}
%\vskip -0.2in
%\begin{small}
\begin{center}
    \begin{tabularx}{0.8\textwidth} { 
       >{\centering\arraybackslash}X
       |>{\centering\arraybackslash}X 
       >{\centering\arraybackslash}X
       >{\centering\arraybackslash}X  }
     %\hline 
     Network & Name & Type & Parameters \\ 
     \hline
   $\beta$MVAE & encoder  &  ResNet18 & 11.2 M \\ 
    & decoder &  ResNet18 &  8.6 M \\
    & outc & OutConv &  5.8 K  \\
    & $\mu'$ & Linear &  128 K  \\
    & $\Sigma'$ & Linear &  128 K  \\
    \hline
    $\beta$MVUnet & encoder  &  ResNet18$^\dagger$ & 286.84 M \\ 
    & decoder &  ResNet18$^\dagger$ &  28.05 M \\
    & outc & OutConv &  4.4 K  \\
    & $\mu'$ & Linear &  256 K  \\
    & $\Sigma'$ & Linear &  256 K  \\
    %\hline
    \end{tabularx}
\end{center}
%\end{small}
%\vskip -0.2in
\end{table}

\begin{figure}[!t]
\centering
%\vskip -0.2in
\includegraphics[width=\textwidth]{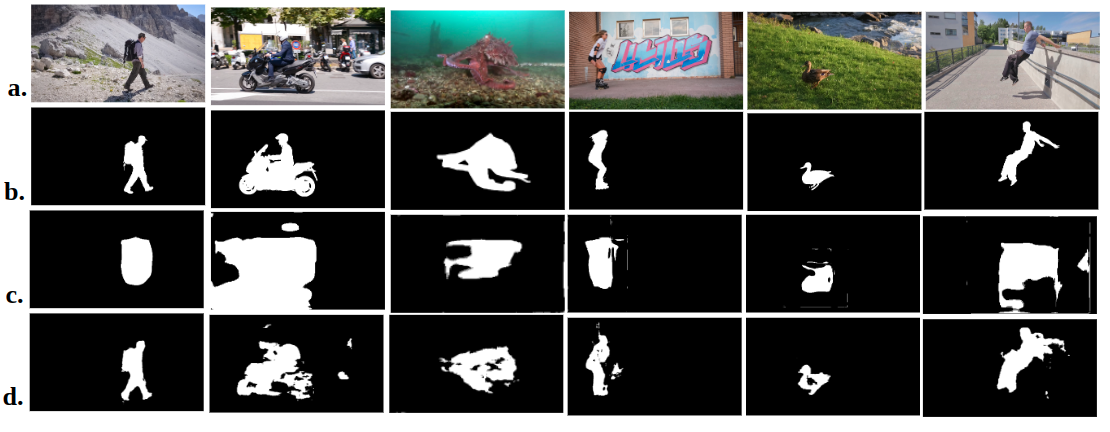}
\caption{The qualitative effect of the skipping connections and one extra convolution layer on the performance. a) frame, b) annotation, c)$\beta$MVAE, d)$\beta$MVUnet.}
\label{abl}
%\vskip -0.2in
\end{figure}

%, as shown Figure \ref{MVUnet}
Now that our method is primarily developed via an auto-encoder network, we apply this approach in another neural network architecture. Since predicting a segmentation mask directly from multivariational inferences is a complicated task for the decoder, we connect the corresponding layers of the encoder to the decoder to build a U-Net. The structure of U-Net is widely employed for the segmentation task in previous works \cite{baheti_eff-unet_2020},~\cite{li_h-denseunet_2018},~\cite{weng_nas-unet_2019}. Moreover, a new convolutional layer is connected to the encoder and the decoder to increase the trainable parameters. The higher number of the trainable parameters facilitates the complex tasks learning for the network. After applying these modifications, the encoder and the decoder each have four main convolutional layers. We named this new network $\beta$MultiVariational UNet ($\beta$MVUnet) trained the same way as $\beta$MVAE, Equation \ref{eq1}, following the algorithm steps of Algorithm \ref{alg:1}. Moreover, the other settings, such as learning rate scheduler, batch size and $\beta$ are unchanged. Table~\ref{Arch} presents characteristics of the $\beta$MVAE and $\beta$MVUnet networks. The $\beta$MVUnet has $277 M$ trainable parameters overall compared to the number of trainable parameters, $20M$, for $\beta$MVAE. Although the network parameters and sizes are increased in the new neural network, the obtained segmentation masks appear to be closer to the annotated ground-truth (see Figure \ref{abl}). Further information and assessments are discussed in Section~\ref{Results} and Appendix.
%The training period of $\beta$MVUnet was approximately $12$ days for $280$ epochs through a PC with $8$ CPUs and one GPU (RTX 3090).

%The learning rate was changed from .... to ... with RAdam scheduler.   

%For BmvUnet that is trained on CC, b is set to b=4!! 

%------------------------------------------------------------------------- 

\section{Results} 
\label{Results}
%We evaluate our networks in terms of three main tasks containing posterior and likelihood estimation, video object segmentation and saliency detection. 

%In the following subsections, the experimental results are explained in particular.  

\subsection{Posterior and log-likelihood evaluation}

\begin{table*}[t]
\caption{The Mahalanobis distance ($\mathcal{D}_{Mah}$), Negative Log-Likelihood (${NLL}$) and Mean Square Error ($MSE$) obtained over our training set, validation set (sequences of YouTubeVOS dataset), and test set (DAVIS16~\cite{perazzi_benchmark_2016}) to assess the encoder and decoder performance in estimating the posterior and likelihood.}
\label{DMah}
%\begin{small}
\begin{center}
    \begin{tabularx}{0.96\textwidth} { 
       >{\centering\arraybackslash}X
       |>{\centering\arraybackslash}X 
       |>{\centering\arraybackslash}X
       >{\centering\arraybackslash}X
       >{\centering\arraybackslash}X  }
     %\hline 
     Method & Data & $\mathcal{D}_{Mah}$ & ${NLL}$ & $MSE$ \\ 
     \hline
   $\beta$MVAE & training set  &  $0.10$ & $4.21$ & $0.71$  \\ 
    & validation set &  $0.05$ & $5.32$ & $0.64$ \\
    
    & test set & $0.06$ & - & $0.65$ \\
    \hline
   $\beta$MVUnet & training set & $1.05 \times e^{-6}$ & $1.98$ & $0.30$  \\ 
    & validation set & $1.22 \times e ^{-6}$ & $2.71$ & $0.30$ \\
    & test set & $0.47 \times e^{-6}$ & - & $0.24$ \\
     %\hline
    \end{tabularx}
\end{center}
%\end{small}
\end{table*}

\textbf{Dataset $\&$ Metrics.} The DAVIS16~\cite{perazzi_benchmark_2016} is used to assess the networks for posterior and likelihood estimation as the test set. It contains $50$ video sequences with pixel-annotated masks. The Mahalanobis distance, which computes the probability distribution distances, is used as a metric to compute the closeness of the prior and the posterior as follows:

\begin{equation}
    \mathcal{D}_{Mah} = (\mu - \mu') \times (\frac{\Sigma + \Sigma'}{2})^{-1} \times (\mu - \mu') 
\end{equation}

\noindent where $N(\mu, \Sigma)$ and $N(\mu', \Sigma')$ are the prior and the posterior, respectively. The Negative Log-Likelihood (NLL) and Mean Square Error (MSE) evaluate the estimated masks by the networks.

\textbf{Evaluation.} As shown in Table  \ref{DMah}, we calculated the Mahalanobis distance across the training, validation, and test sets. The computed distances show the ability of our networks to mimic the prior using different datasets. The NLL and MSE scores, on the other hand, evaluate network generalisation and overall error in segmentation masks. The NLL, as explained by the authors of VQ-VAE2~\cite{razavi_generating_2019}, is also a criterion for validating the overfitting problem in training. Because learning object segmentation for a large dataset is a difficult task, the difference in NLLs between the training and validation sets, which is approximately $0.7$, is a reasonable value. The realisation that our encoder and decoder did not overfit for the $\beta$MVUnet method is reflected in the fact that the MSEs for both training and validation sets were the same.

In addition to the larger MSE, we obtained a greater distance between the prior and the posterior for $\beta$MVAE. This result is also realistic because the architectural difference influences the outcomes. In terms of size and parameters, the $\beta$MVUnet is larger. It is also aided by the skipped connections. These characteristics improve network performance. Similarly, the NLL of the $\beta$MVUnet is lower than that of the $\beta$MAVE. The presence of NLLs in the training and validation sets indicates that the encoder and decoder are not overfitted for $\beta$MAVE. We conducted an experiment to add the Planar, Radial~\cite{rezende_variational_2015}, and RealNVP flows~\cite{dinh_density_2017} at the bottleneck of our networks and learn the fully factorised version of our prior in order to compare our networks with the literature in terms of posterior estimation task. However, we encountered posterior collapse~\cite{lucas_understanding_2019} over our training data. The massive data handled in our approach, we believe, is the underlying cause of these methods failing. The IAF method~\cite{kingma_improving_2017} was also investigated and found to have the same problem, namely posterior collapse.

\subsection{Video object segmentation}

\textbf{Dataset $\&$ Metrics.} We assessed the performance of our proposed networks in segmentation task using DAVIS16 dataset. Following the DAVIS16 protocol, the region similarity (Jaccard index $\mathcal{J}$) and contour similarity $\mathcal{F}$ are employed as the metrics. 

\begin{figure}[!t]
\vskip -0.2in
\centering
\includegraphics[width=0.8\textwidth]{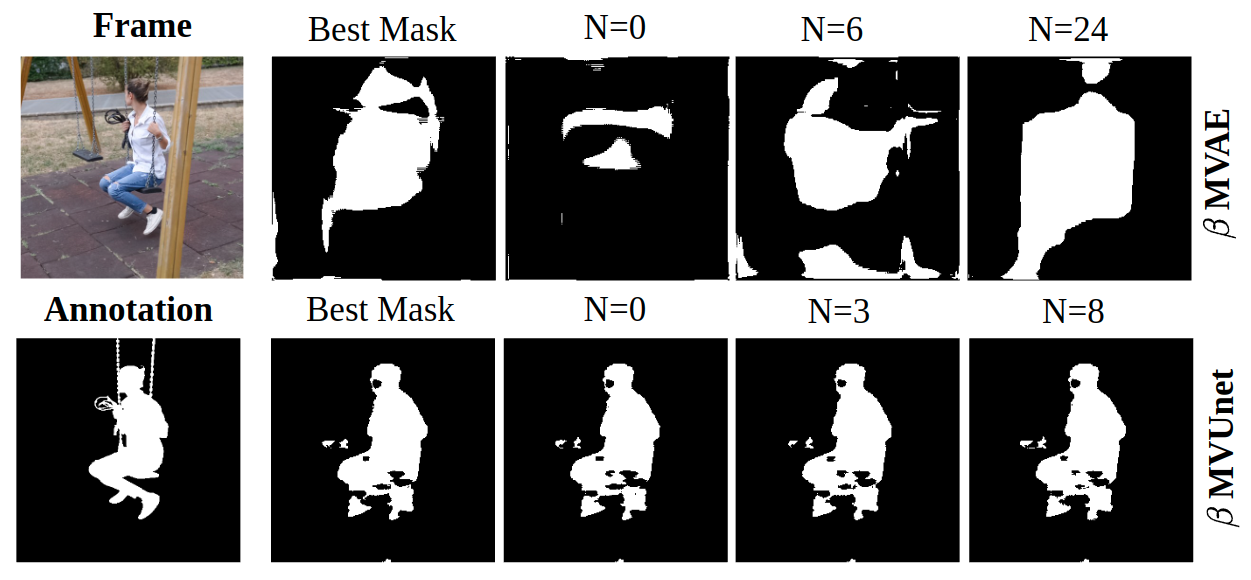}
\caption{The stochastic feature of the bottleneck affects the video segmentation results. For one single frame, the created masks are different depending on their layer place $N$ in the input batch.}
\label{rmsk}
\vskip -0.2in
\end{figure}

\noindent \textbf{Evaluation.} As is explained earlier, our method applies a preprocessing step on the frames to create frame patches and feed them to the networks. Subsequently, we have to resize and replace the output of the network in a zero mask to obtain the binary object mask. Another aspect of our segmentation performance concerns the role of the stochastic feature of the bottleneck. By tiling one single image in batch size $32$, we formed an appropriate size of the input for our networks. Subsequently, we have $32$ different binary masks which are not exactly the same due to random sampling in the bottleneck. In Figure \ref{rmsk}, several resulting masks for one single frame patch are shown according to their layer number $N$ in the batch. The best mask is the one which has the greatest region similarity with the annotation among the $32$ estimated maps. For the $\beta$MVUnet network, the computed maps are more alike than for $\beta$MVAE. This demonstrates the impact of skipping connections from the encoder layers to the decoder that diminish the stochastic role of the bottleneck in the $\beta$MVUnet decoder's outcome.

\begin{figure*}[!t]
%\vskip -0.2in
\centering
\includegraphics[width=\textwidth]{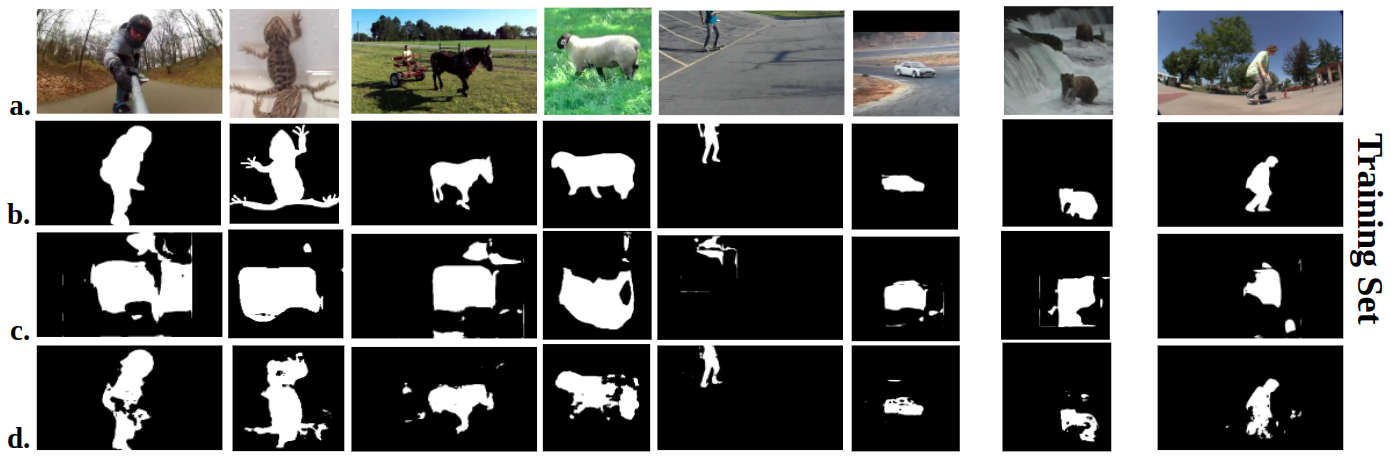}
\includegraphics[width=\textwidth]{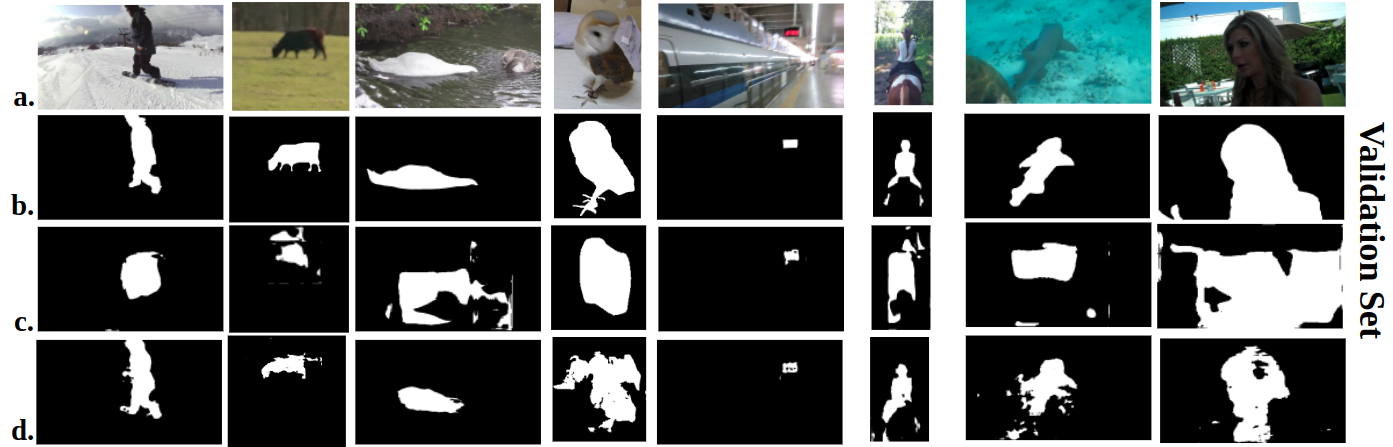}
\includegraphics[width=\textwidth]{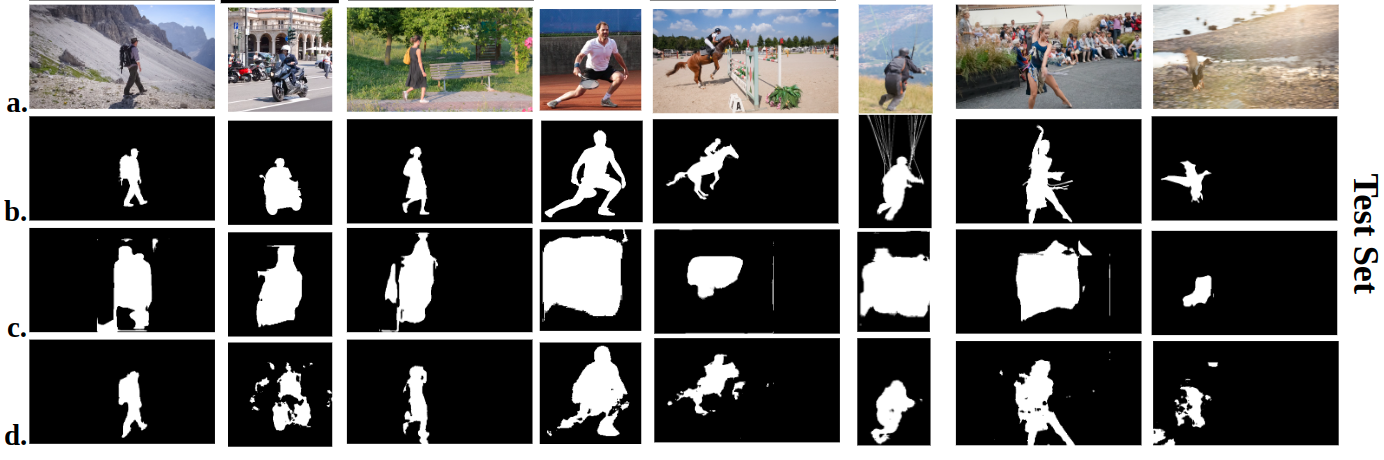}
\caption{Video segmentation for several frames of training (YouTube-VOS), validation (YouTube-VOS), and test (DAVIS16) sets. a) input frame, b) annotation, c) $\beta$MVAE result, d)$\beta$MVUnet output.}
\label{segMapVid}
%\vskip -0.2in
\end{figure*}

 Figure \ref{segMapVid} exemplifies several masks generated by our networks in comparison to the annotations added to the training, validation and test set. For some classes such as people, the segmentation maps of the $\beta$MVUnet represent more accurate masks (e.g. first column in the training. validation and test sets, eighth column in the training and validation set). For large objects, the $\beta$MVAE functions better such as for vehicles (e.g. second column in the test set). Broadly, we can state that $\beta$MVUnet constitutes binary masks with more precise boundaries than $\beta$MVAE. This observation is confirmed with the computed metrics, as indicated in Table \ref{Davis16}. The contour similarity $\mathcal{F}$ over DAVIS16 is $0.43$ for $\beta$MVUnet, while this metric is $0.18$ for $\beta$MVAE network. This emphasizes that the $\beta$MVUnet elaborates the binary masks with fastidious boundaries in comparison to the $\beta$MVAE network. Moreover, our networks is trained to learn two different tasks simultaneously. Therefore, the segmentation performance of our method is fair to evaluate with similar methods trained for multi-tasks, without using temporal information, without using memory, and processed in one single scale. The SAGE~\cite{wang_saliency-aware_2015} method satisfies the last three conditions, while it does not develop based on deep neural networks. Based solely on a single frame patch, we constructed the segmentation masks from the neural network with a quality almost equal to that of the SAGE~\cite{wang_saliency-aware_2015} method(see Table \ref{Davis16}). For the methods only trained for the video object segmentation, e.g. by SegFlow~\cite{cheng_segflow_2017}, higher numbers are reported for $\mathcal{F}$ and $\mathcal{J}$. This method employed a two-branches network, including ResNet101 which is a larger network than ours (ResNet18). Also, SegFlow used the optical flows to perform video object segmentation.

\begin{table*}[!t]
\caption{Object segmentation performance on the DAVIS16 dataset~\cite{perazzi_benchmark_2016} in terms of region similarity $\mathcal{J}$ and contour similarity $\mathcal{F}$. The \textbf{ResNet18$^\dagger$} is the ResNet18 architecture plus the one extra convolutional layer and OF is Optical Flow.}
\label{Davis16}
%\begin{small}
\begin{center}
    \begin{tabularx}{\textwidth} { 
       >{\centering\arraybackslash}X
       |>{\centering\arraybackslash}X 
       >{\centering\arraybackslash}X
       >{\centering\arraybackslash}X
       >{\centering\arraybackslash}X
       >{\centering\arraybackslash}X}
     %\hline 
     Methods & Backbone & OF & $\mathcal{J} \& \mathcal{F}$ & $\mathcal{J} \uparrow$  & $\mathcal{F} \uparrow$ \\ 
     \hline
     SegFlow~\cite{cheng_segflow_2017} & ResNet101 & \checkmark & $67$ & $67.4$ & $66.7$ \\ 
     SAGE~\cite{wang_saliency-aware_2015} & - & \checkmark & $40.4$ & $42.6$ & $38.3$ \\ 
     %ARP~\cite{koh_primary_2017} & $\times$ & \checkmark & $73.6$ & $76.2$ & $71.1$ \\ 
     %CIS~\cite{yang_unsupervised_2019} & \checkmark & \checkmark & $71$ & $71.5$ & $70.5$ \\ 
     %\hline
    {$\beta$MVAE} & ResNet18 & $\times$ & $28.3$ & $38.4$ & $18.3$ \\ 
    %\hline
     {$\beta$MVUnet} & ResNet18$^\dagger$ & $\times$ & $41.6$ & $40.4$ & $42.9$ \\ 
     %\hline
    \end{tabularx}
\end{center}
%\end{small}
\end{table*}

\begin{figure}[!ht]
\centering
\vskip -0.25in
\includegraphics[width=0.96\textwidth]{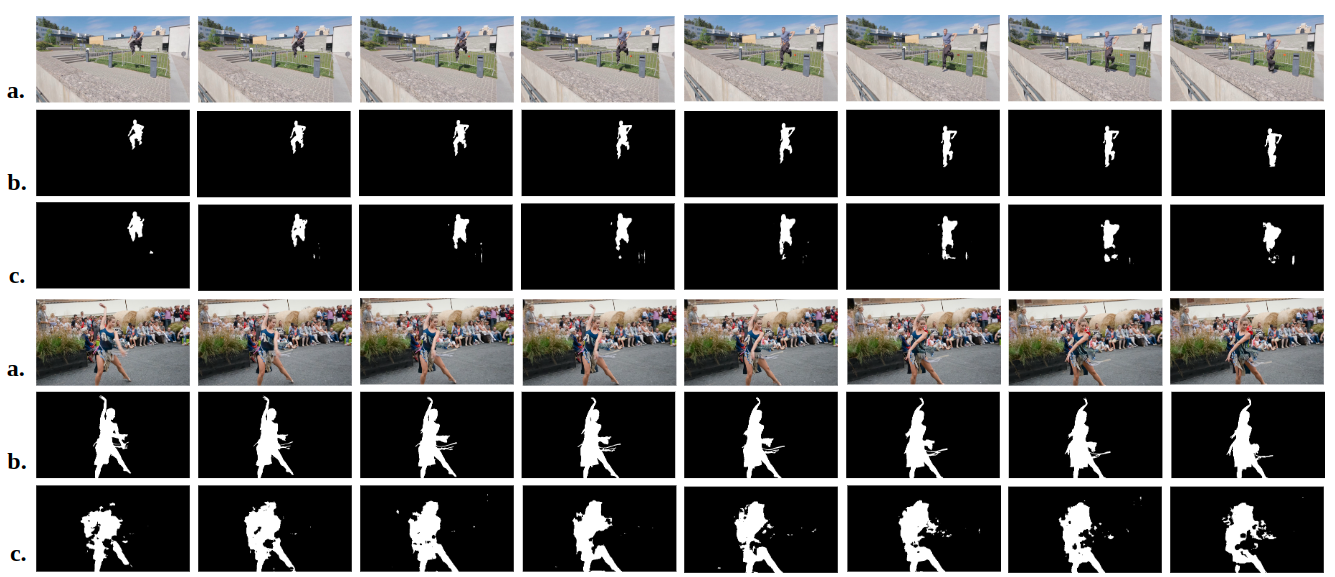}
\includegraphics[width=0.96\textwidth]{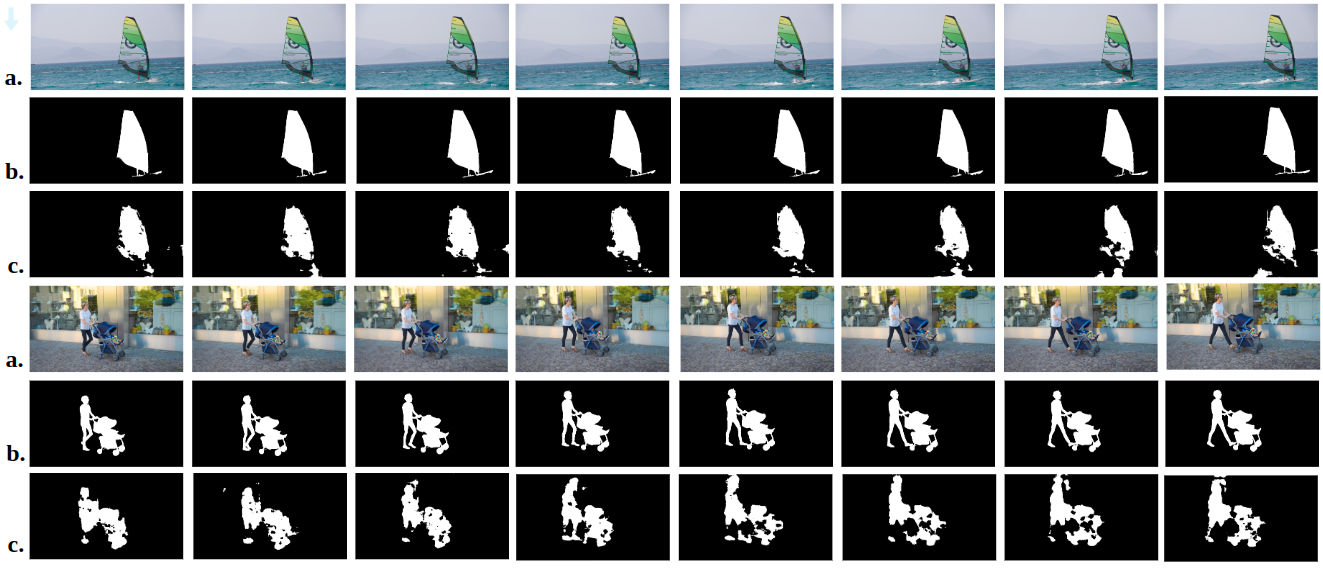}
\includegraphics[width=0.96\textwidth]{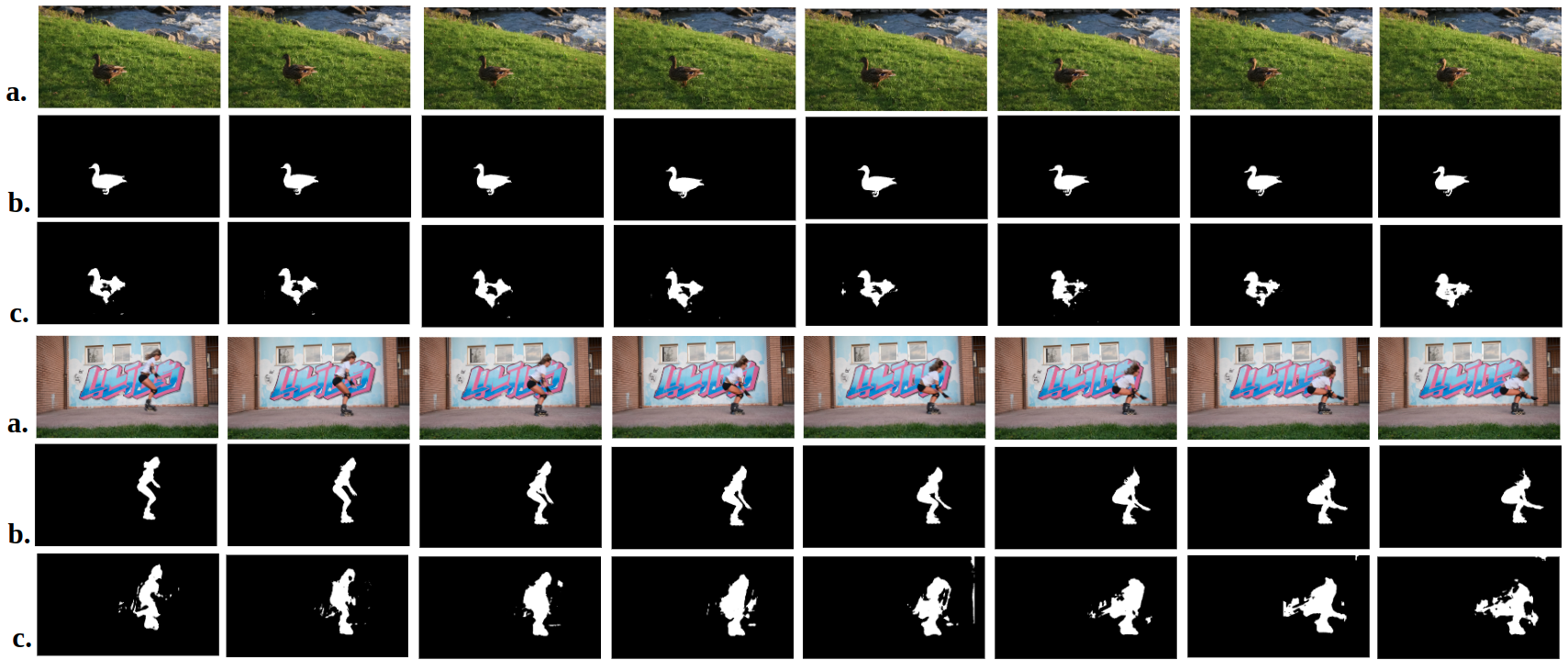}
\caption{Several sequences of DAVIS16 dataset to indicate the binary mask for successive time steps (Left to right). a) frame, b) annotation, c)$\beta$MVUnet.}
\label{vidSeq}
\vskip -0.25in
\end{figure}

We also illustrate our network's performance across frame to visualize the time consistency in object mask estimation, see Figure \ref{vidSeq}. For four video sequences of the Davis16 dataset, the binary mask estimated by $\beta$MVUnet is indicated. Despite the cluttered background, different human poses (such as dancing in second sequence, walking in the third sequence, and rollerblading in the last sequence) the moving object is detected. The video frames are selected from diverse object categories including human, animal, and objects.

\subsection{Saliency detection}

\textbf{Dataset $\&$ Metric.} The SED2~\cite{ge_image-segmentation_2006} and ECSSD~\cite{shi_hierarchical_2016} datasets are employed for saliency detection assessments in terms of MAE and F-measure metrics. These datasets contain $100$ images in SED2 and $1000$ images in ECSSD and their corresponding annotations. The images of SED2 contain two salient objects to assess the ability of object recognition where the salient object is not unique in the images and not localized at the center. The ECSSD feature is the cluttered backgrounds. We computed the Mean Square Error (MSE) and F-measure. For F-measure, the coefficient is $0.3$, similar to what is suggested in the literature.

\noindent \textbf{Evaluation.} Delivering the probabilistic map per class instead of the segmentation map, we compute the saliency maps. Since our networks process patches of a size of $256 \times 256$, the image dimension is reduced to obtain the results. Then, the output is resized in the real size of the image. This results in blurry boundaries and some cases, details are lost. Similar to the previous experiment, we obtain the best mask per input but instead of the mask, the probabilistic map corresponding to the best mask is reported as the output.

\begin{figure*}[!ht]
\vskip -0.2in
\centering
\includegraphics[width=\linewidth]{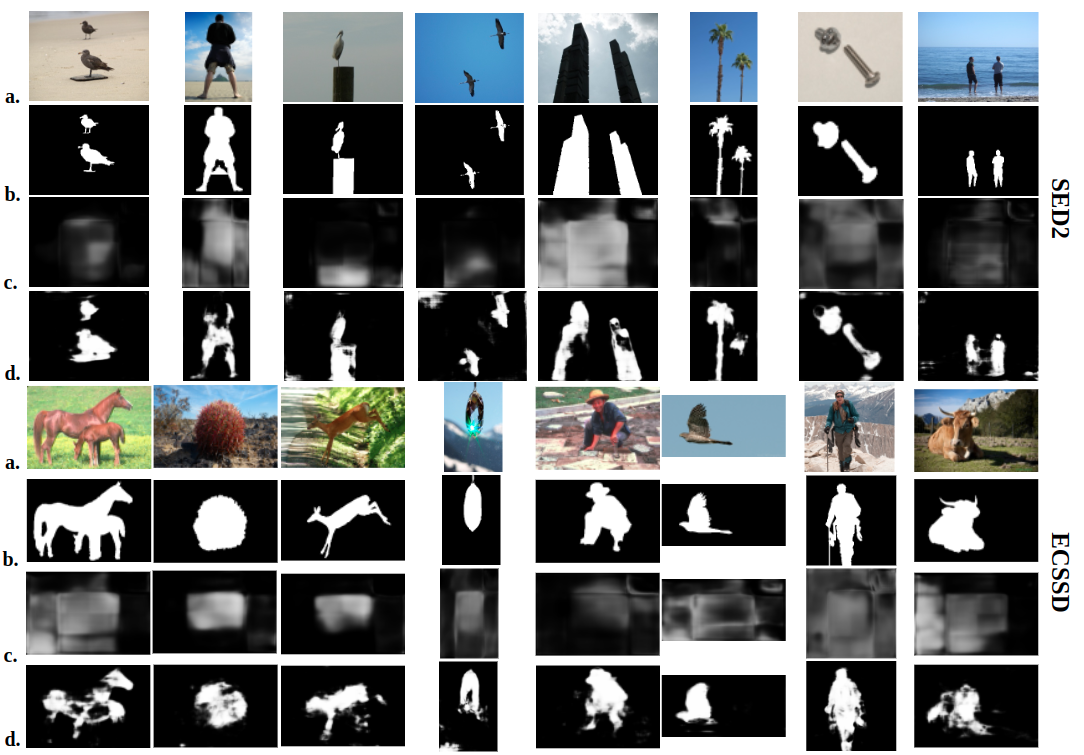}
\caption{The visualized probabilistic maps as saliency maps for SED2 and ECSSD datasets. a)Input Image, b)Annotation, c)$\beta$MVAE map, d)$\beta$ MVUnet map.}
\label{sal}
\vskip -0.2in
\end{figure*}

In Figure \ref{sal}, we show the saliency map generated by $\beta$MVAE and $\beta$MVUnet for saliency detection datasets. Although we trained our network with object-centred images, $\beta$MVUnet detects both objects of SED2. From animals to unknown objects, $\beta$MVUnet segments the salient object in different backgrounds (columns 2 and 4 in Figure \ref{sal}). Table \ref{Ecssd} compares the saliency detection ability of our method with that of several other methods. The 3Graph~\cite{nouri_salient_2018} processed spatial features of the images to create three graphs based on the contrast values of the pixels. We obtain better scores compared to 3Graphs. However, for the other method DeepSal~\cite{li_deepsaliency_2016}, both metrics are better. The DeepSal~\cite{li_deepsaliency_2016} employed the pre-trained VGG16 network to perform multi-task learning including the saliency detection task. DeepSal also has a refinement branch containing a regularized regression that works on a superpixel graph of the image. Specially trained for the task, greater size for the network and refinement branch are the main reasons for the DeepSal to outperform our networks. 

%Considering all of these advantages, the DeepSal 

%The state-of-the-art methods in salient object detection outperform our networks. However, these methods~\cite{liu_simple_2019}, ~\cite{wei_fnet_2020} utilized multi-scale processing with sizable different networks, e.g. ResNet-50~\cite{he_deep_2016}. Furthermore, these networks are specialized only for segmentation tasks whereas our methods overcame the complex posterior estimation over a large dataset in addition to the segmentation learning.

\begin{table}[t!]
\caption{Saliency detection results over the ECSSD dataset in terms of Mean Absolut Error(MAE) and F-Measure(FM). The \textbf{ResNet18$^\dagger$} is the ResNet18 architecture plus one extra convolutional layer.}
\label{Ecssd}
%\begin{small}
\begin{center}
    \begin{tabularx}{\textwidth} { 
       >{\centering\arraybackslash}X
       |>{\centering\arraybackslash}X 
       >{\centering\arraybackslash}X
       >{\centering\arraybackslash}X }
     Methods & Backbone & MAE $\downarrow$ & FM $\uparrow$  \\ [1ex]
     \hline
     3Graphs~\cite{nouri_salient_2018} & - &  $0.20$ & $0.42$  \\ [1ex]
     %PoolNet~\cite{liu_simple_2019} & ResNet50 &  $0.04$ & $0.94$  \\ [1ex]
     DeepSal~\cite{li_deepsaliency_2016} & VGG16 &  $0.16$ & $0.75$  \\ [1ex]
    {$\beta$MVAE(ours)} & ResNet18  & $0.20$ & $0.39$  \\ [1ex]
    %\hline
     {$\beta$MVUnet(ours)} & ReseNet18$^\dagger$ & $0.19$ & $0.50$  \\ [1ex]
    % \hline
    \end{tabularx}
\end{center}
%\end{small}
\end{table}
%\begin{table}[t]
%\caption{Salinecy detection over SED2 dataset.}
%\label{Davis16}
%\begin{small}
%\begin{center}
%   \begin{tabularx}{0.6\textwidth} { 
    %   |>{\centering\arraybackslash}X
    %   |>{\centering\arraybackslash}X 
    %   >{\centering\arraybackslash}X
    %   >{\centering\arraybackslash}X| }
%     \hline 
    % Methods & Deep NN & MAE & FM  \\ 
    % \hline
    %3Graphs~\cite{nouri_salient_2018} & $\times$ &  $0.14$ & $0.69$  \\ 
    % SAGE~\cite{wang_saliency-aware_2015} & $\times$ &  $40.4$ & $42.6$  \\ 
    % ARP~\cite{koh_primary_2017} & $\times$ &$73.6$ & $76.2$  \\ 
    % CIS~\cite{yang_unsupervised_2019} &  \checkmark & $71$ & $71.5$ \\ 
%     \hline
%    \textbf{$\beta$MVAE} & \checkmark  & $0.20$ & $0.30$  \\ 
    %\hline
%     \textbf{$\beta$MVUnet} & \checkmark & $0.18$ & $0.49$  \\ 
%     \hline
%    \end{tabularx}
%\end{center}
%\end{small}
%\end{table}

%\section{Discussion}

%\subsection{Limitations}

%In terms of technical views, our implementation struggles with GPU and CPU interaction in the bottleneck which leads us to a slower training process. Our estimated parameters have to be sent by $\mu'$ and $\Sigma'$ into the CPU for the covariance matrix estimation accomplished by a Numpy~\cite{harris_array_2020} function, \texttt{np.conv}. Concerning outcomes of the trained networks, the segmentation results are less satisfactory compared to the results of other video segmentation methods. Note that they used temporal information such as optical flow~\cite{cheng_segflow_2017} or applied multi-scale post-processing~\cite{yang_unsupervised_2019} to refine the generated masks. 

\section{Conclusion and Future works}

Entangled representation is the task of learning the relations between the variables. It is an essential task in computer vision to model asymmetric systems and notions of real-world problems. We proposed an asymmetric idea to model a single object's motion across video frames. Data analysis on our motion parameters represented a multivariate Gaussian distribution over parameters. We formulated $\beta$MVAE to learn this distribution plus the segmentation mask of the moving object. We demonstrated that our proposed method properly learns the KL divergence of the estimated posterior and the prior. The decoder of $\beta$MVAE estimates the binary mask of the object in the input frame. The latent variables in our setting correspond to a set of tracking parameters derived from our scenario of motion modelling in the video frames. These variables are dependent on each other and describe the single object's motion. Their relation is formulated as a multi-variate Gaussian distribution by a mean vector and a covariance matrix. Our proposed $\beta$MVAE succeeded in learning the non-diagonal full covariance matrix in addition to the segmentation features of the moving object. We also improved the results using a UNet structure trained using our approach. The second network, $\beta$MVUnet surpassed the $\beta$MVAE in both posterior estimation and segmentation mask creation.

%\section{}

We aim to extend our work into a single object tracking method. We are going to include a specific network to the current $\beta$MVUnet to predict the motion parameters of the moving object in the video frames. As we plan to learn the tracking task with the soft actor-critic algorithm, we proposed $\beta$MVAE to introduce into our approach of learning an MGD from a mean vector and covariance matrix. Predicting the covariance matrix is vital to precisely mapping video patches into the latent space where each dimension is related to one parameter of the tracking system. The relation of every two variables is deciphered with the covariance values. The next step of our research is to modify the SAC algorithm to learn the tracking task while the action set (tracking parameters) is selected from our target latent space.





\section{Acknowledgements}

The National Sciences and Engineering Research Council of Canada(NSERC) provided funding for this research through a Discovery Grant (RGPIN-2016-05876). Also, the authors would like to thank the Fonds de Recherche du Québec, Nature et Technologie (FRQNT) for financial support under grant RS4-265455 - REPARTI.

The authors appreciate the discussion and critical comments of Setareh Rezaee. We would additionally like to thank Annette Schwerdtfeger for proofreading the paper.

\bibliographystyle{elsarticle-num}
\bibliography{main.bib}

%\begin{thebibliography}{00}


%\bibitem{}

%\end{thebibliography}

\clearpage

\begin{center}
{\Large\bfseries
Appendix} \\~\\ {\large $\beta$-Multivariational Autoencoder for Entangled Representation Learning in Video Frames}
%\vskip .5em
%\normalfont
%{First Last, First Last \& First Last\symfootnote{Both acknowledgments and author affiliation information go in an initial footnote like this, referenced by an asterisk on the end of the last author's name on the author-name line at the top. You can say who you would like to thank here and then end the footnote as follows. Authors: Author One, University of the Atlantic (\href{mailto:author1@atlantic.edu}{author1@atlantic.edu}) \& Other Author, Pacific University (\href{mailto:author2@gmail.com}{author2@gmail.com}).}}
\vskip .5em
\end{center}

%\section{Data Analysis}

\setcounter{section}{0}
\section{Uniqueness of $G_t$ in our motion model}

It is essential to verify if it is feasible to calculate $G_t$ for every two successive frames and whether it is a unique transformation in each time step $t$ or not. 

Let us assume the current and next \textit{rbbox} as:

\begin{align}
    rbbox_t = ([x, y], [s_x, s_y], \alpha) \\
    rbbox_{t+1} = ([x', y'], [s_x', s_y'], \alpha')
\end{align}
    
Following equations are used to calculate the scale in $x$ axis, scale in $y$ axis, rotation, translation in $x$ axis, and the translation in $y$ axis.  

 \begin{align}
     \label{3} & \Delta s_x = \frac{s_x'}{s_x} \\ 
     \label{4} & \Delta s_y = \frac{s_y'}{s_y} \\ 
     \label{5} & \theta = \alpha' - \alpha \\ 
     \label{6} & t_x  = x' - x \times \Delta s_x \times cos \theta + y \times \Delta s_y \times sin \theta \\ 
     \label{7} & t_y = y' - y \times \Delta s_x \times sin \theta - y \times \Delta s_y \times cos \theta 
 \end{align}
 
 Therefore, there is one unique $G_t$ for every two desirable \textit{rbbox}s. To normalize the translation parameters, we divided the translations parameters by the image dimensions as follows
 
 \begin{align}
     \Delta x = \frac{t_x}{w_I}, \Delta y = \frac{t_y}{h_I}
 \end{align}
 
 where $w_I$ and $h_I$ refer to the image's width and height. Also, we considered the $sin \theta$ for the rotation parameter to scale the numbers.

\clearpage

\section{Data Preparation}
\label{data}

Pixel-wise annotation, the large size of the training set, various objects, and diverse motion patterns make YouTube-VOS~\cite{xu_youtube-vos_2018} one of the largest existing datasets in computer vision. This dataset contains 4453 YouTube video clips annotated pixel-wise for 94 object categories and 78 diverse objects and motions. The training set consists of 3471 video sequences, and the validation set includes 507 video sequences. We used the YouTube-VOS dataset to train our method due to its privileges, including an extensive dataset and diverse video sequences. It is necessary to customize the YouTube-VOS dataset for object tracking since our primary goal was to develop a tracking system. Therefore, a pre-processing step is performed to construct our training and validation data from the training set of the YouTube-VOS dataset.

%\clearpage
\subsection{Pre-processing}
\begin{figure}[!b]
\centering
\includegraphics[scale=0.17]{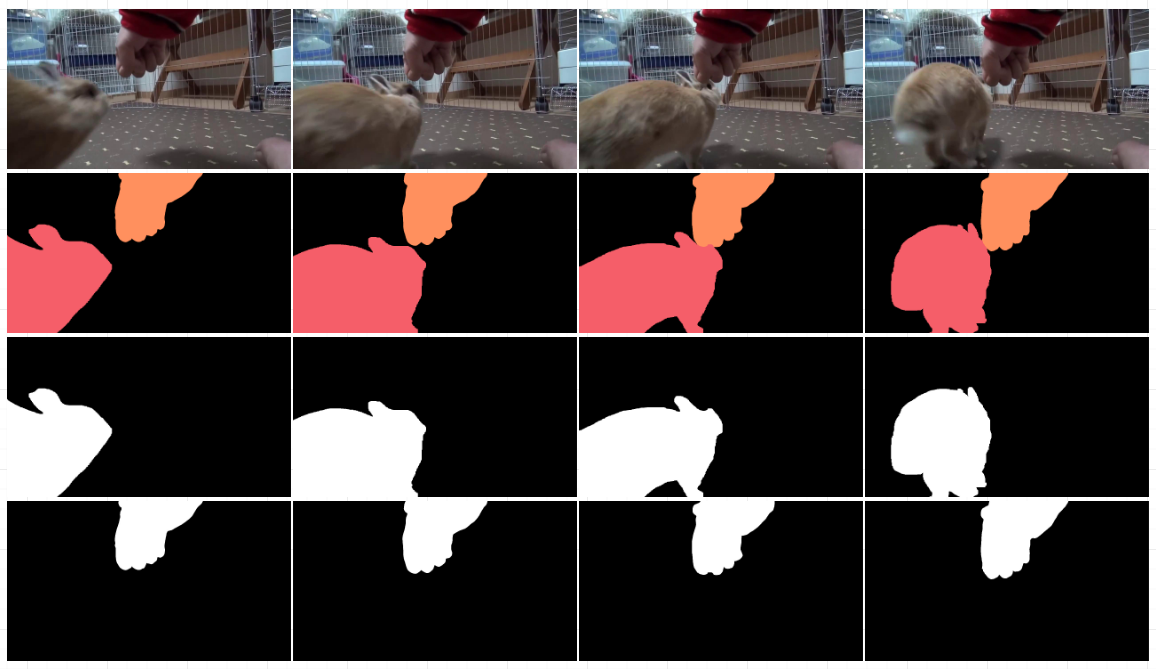}
\caption{Data Preparation Process: The first row shows the video frames, the second row, presents the YouTube-VOS annotation, third and fourth rows demonstrate our binary annotation to create a single object tracking dataset.}
\label{OurData}
\end{figure}

The single object tracking task needs video sequences containing a single moving object while YouTube-VOS is annotated based on multi-object masks. In the primary step, we extracted binary masks out of multi-object masks. Also, we did not consider occlusion in data preparation which means we removed the sequences where the moving object is fully-occluded in at least one frame. Therefore, if one video sequence contains three objects and all of these objects are always present in the frames during the sequence, we generated three video sequences, each related to one object, as shown in Figure \ref{OurData}. 

Furthermore, we eliminated the cases where the object's size is greater that $70\%$ of the image's dimensions. In other words, our dataset contains the videos of a moving object in the frame $w_I \times h_I$ where the object's dimensions $[w, h]$ satisfied these inequalities $w < 0.7 \times w_I$ and  $h < 0.7 \times h_I$. In the end, we created 3105 video sequences with pixel annotations. This data is divided randomly into 2484 sequences for training and 621 sequences for validation purposes.

\subsection{Data analysis}

The training data is composed of video frames preprocessed to be object centered in the size of $256 \times 256 \times 3$ and referred to as frame patches. The ground truth is the cropped mask from the corresponding frame's mask while the object's mask is centered. Using the annotated masks, we calculated the true parameters of $G_t$ for every two successive frames by using Equations~\ref{3} to~\ref{7}. We assumed an MGD over true values of the parameters while the mean vector and covariance matrix are obtained as:

\begin{small}
\begin{flalign}
    \mu &= \begin{bmatrix}
    1.0651, 1.0679, 0.00, 0.0718, 0.083 \end{bmatrix}&& \\ \nonumber
    \Sigma &= \begin{bmatrix}
0.265 & -0.094 & -0.014 & -0.023 & 0.124 \\
-0.094 & 0.332 & 0.011 & 0.056 & -0.054 \\
-0.014 & 0.011 & 0.183 & 0.073 & -0.21 \\
-0.023 & 0.056 & 0.073 & 0.107 & -0.029 \\
0.124 & -0.054 & -0.21 & -0.029 & 0.564 \\
\end{bmatrix} && 
\end{flalign} \label{cov}
\end{small}

\noindent where the elements of $\mu$ and $\Sigma$ correspond to our variables $V = \{ v_1, v_2, v_3, v_4, v_5 \}$, respectively. These variables are in particular the variations of scale in $x$ and $y$ axis, $\Delta s_x$ and $\Delta s_y$, rotation $\theta$, and the variations of translation in $x$ and $y$ axis, $\Delta x$ and $\Delta y$.

\clearpage

\section{Training Curves}

\begin{figure*}[b!]
        \includegraphics[width=0.3\linewidth]{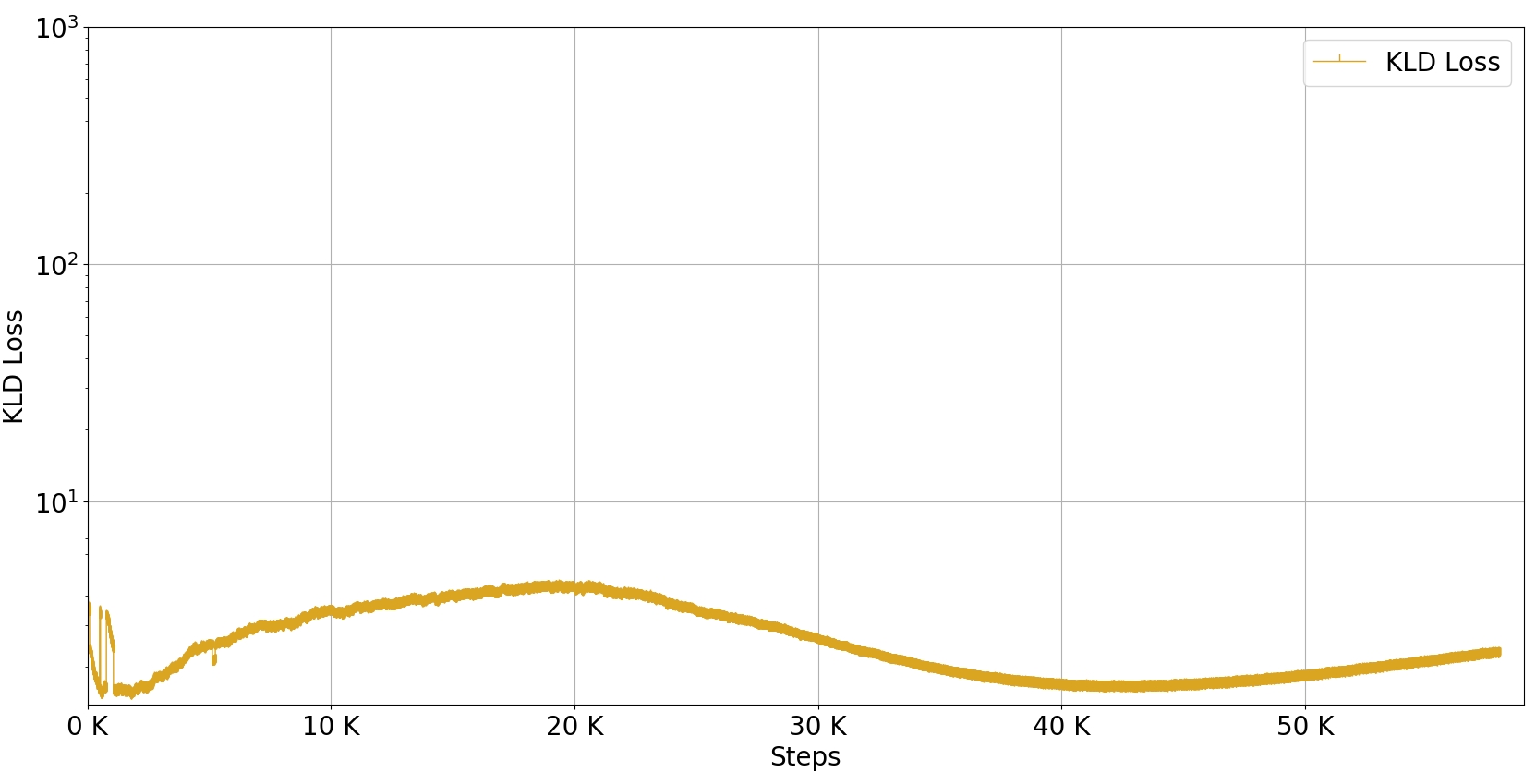}
\hspace*{\fill}
        \includegraphics[width=0.3\linewidth]{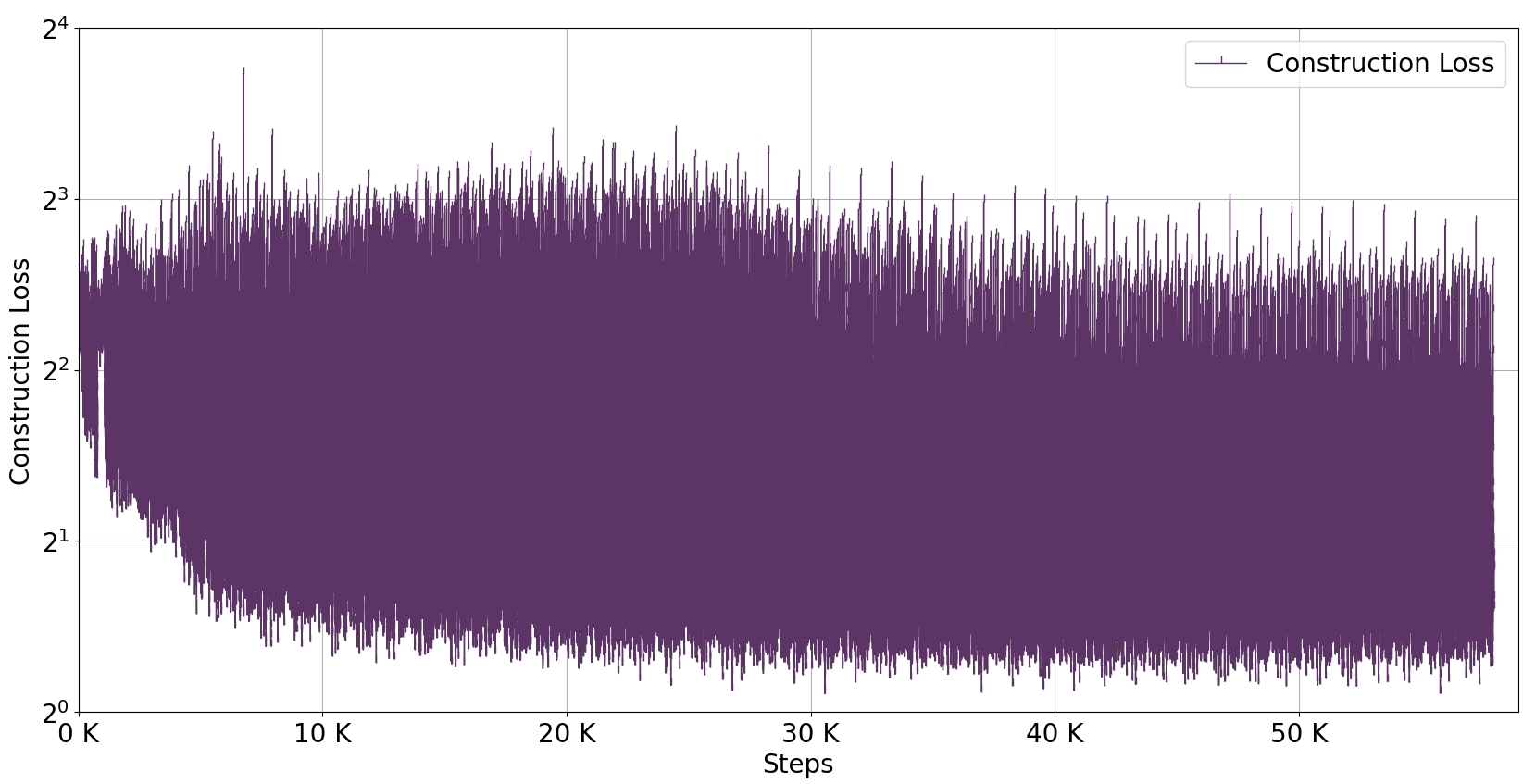}
\hspace*{\fill}
        \includegraphics[width=0.3\linewidth]{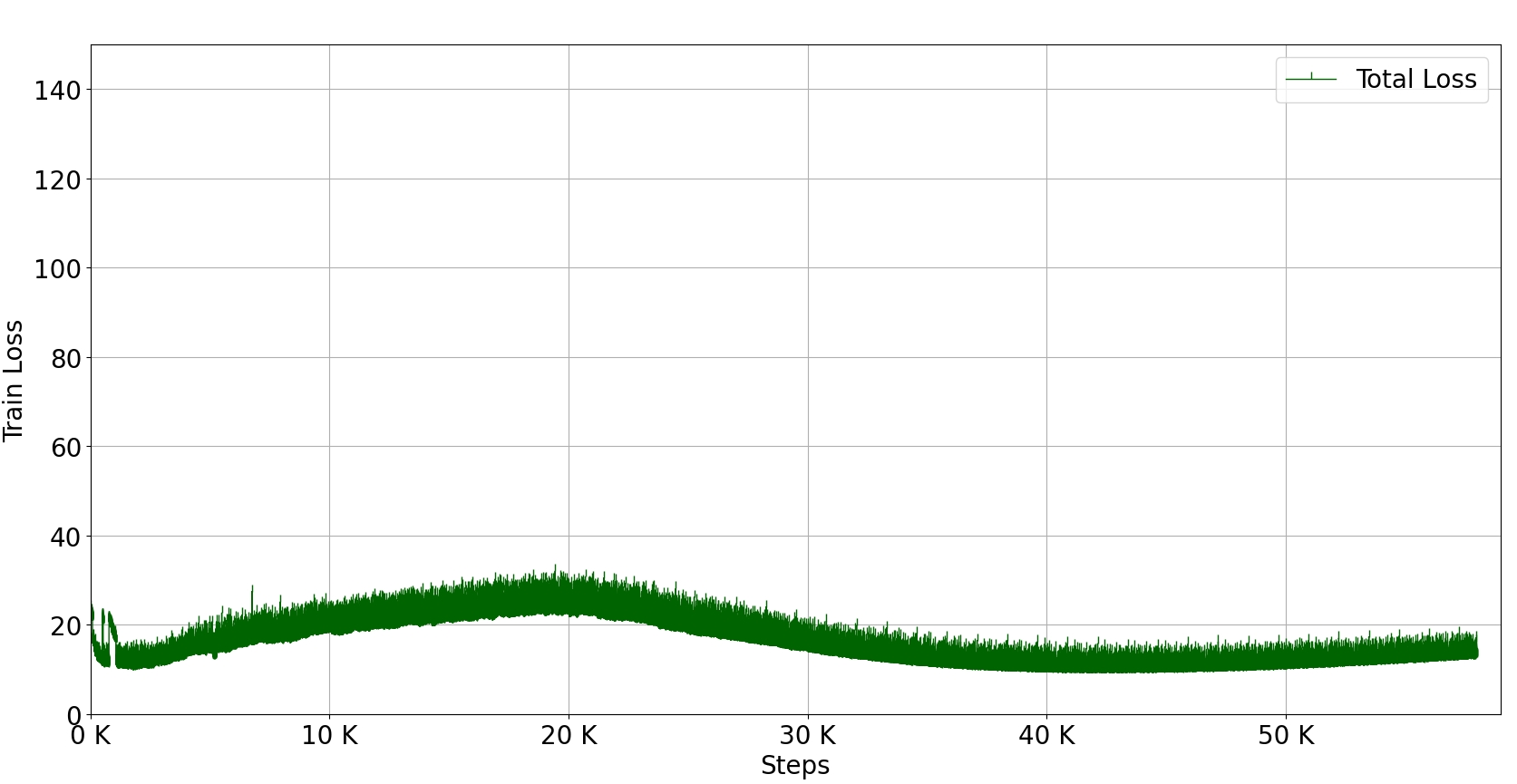}
\caption{The loss curves in the training phase of $\beta$MVAE. Left to right: total loss, construction loss, and KL divergence loss.}
\label{train1}
\end{figure*}

\begin{figure*}[b!]
        \includegraphics[width=0.3\linewidth]{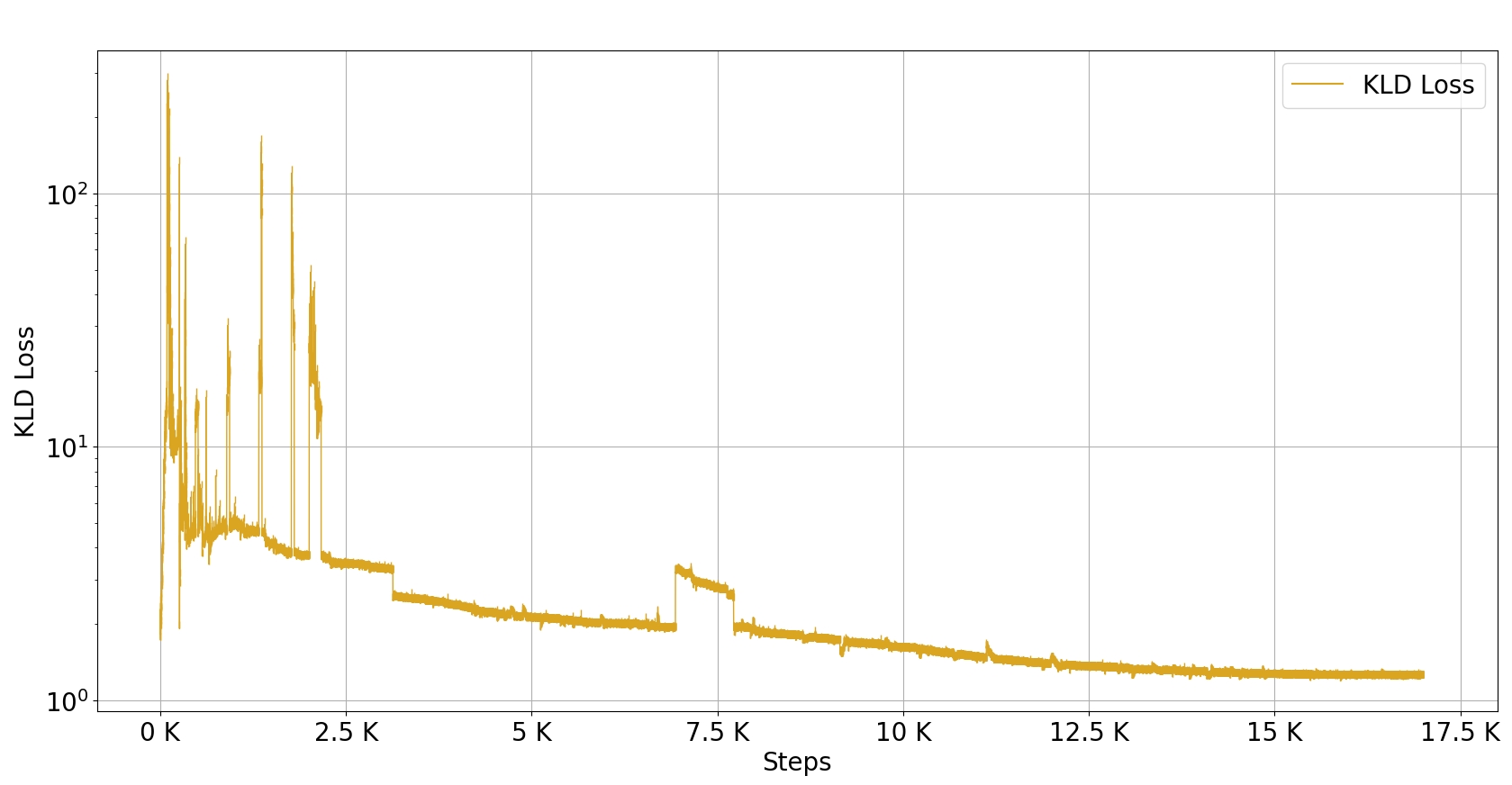}
\hspace*{\fill}
        \includegraphics[width=0.3\linewidth]{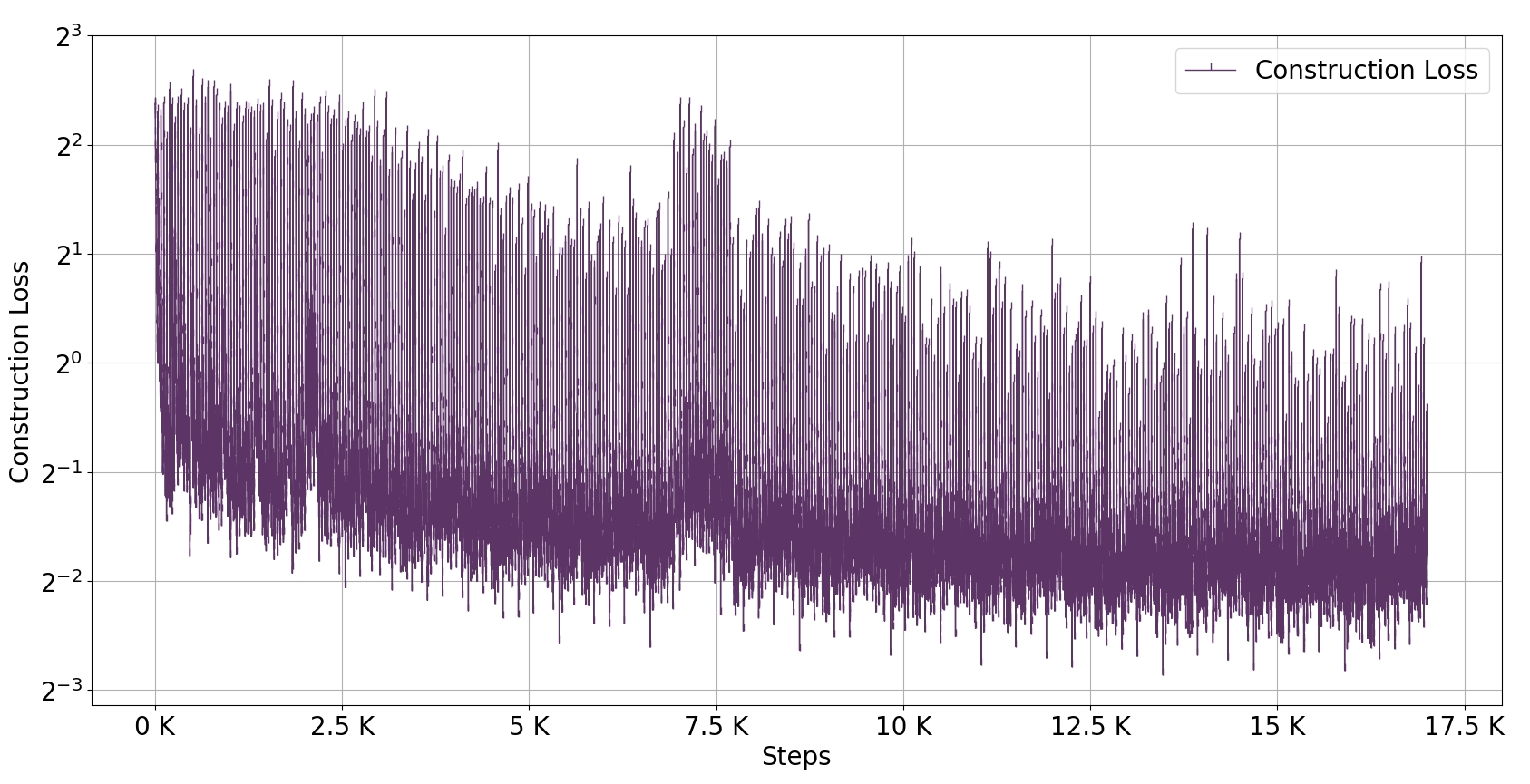}
\hspace*{\fill}
        \includegraphics[width=0.3\linewidth]{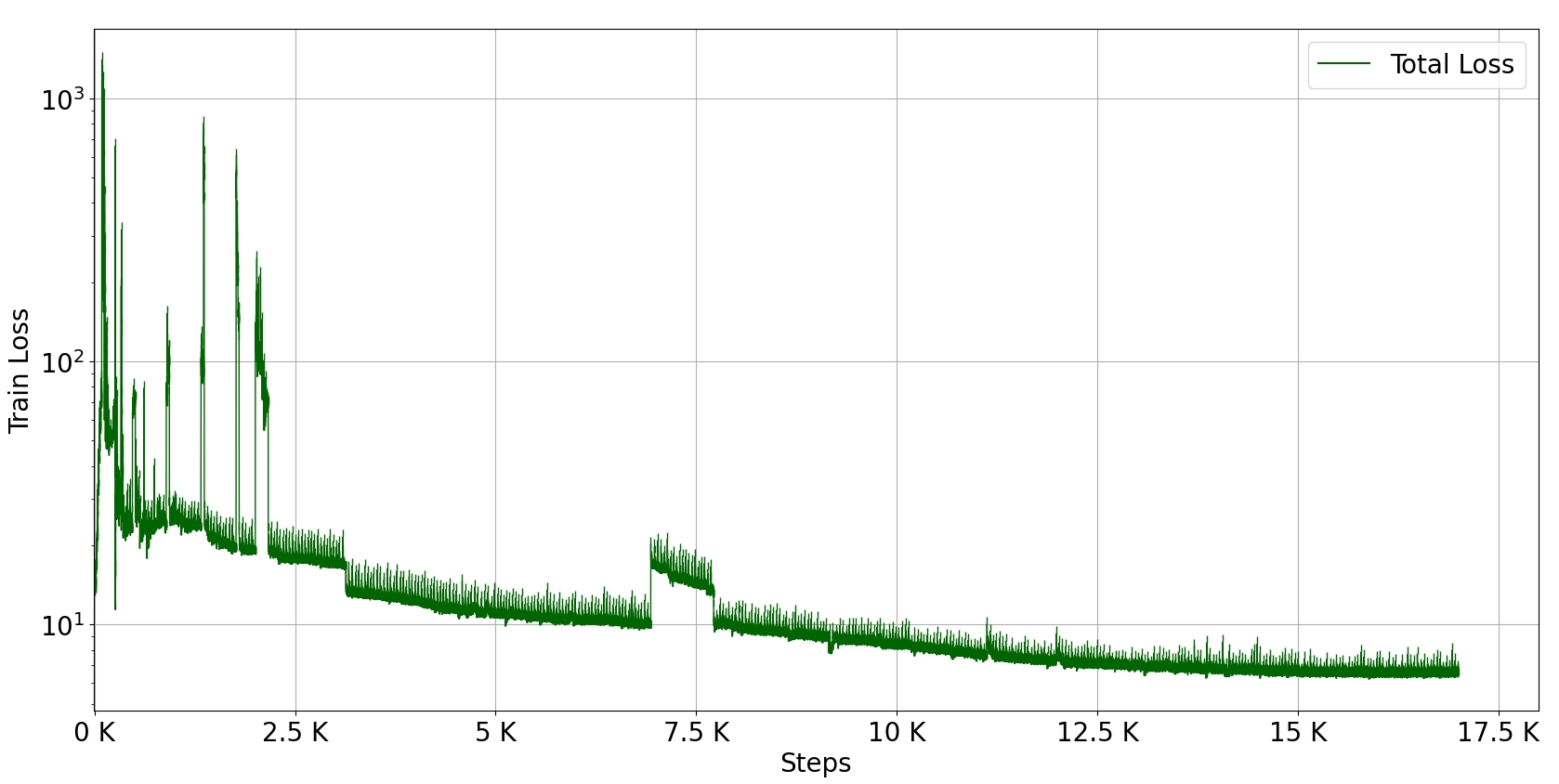}
\caption{The loss curves in the training phase of $\beta$MVUnet. Left to right: total loss, construction loss, and KL divergence loss.}
\label{train2}
\end{figure*}

Figure \ref{train1} illustrates the loss function curves during the training for the $\beta$MVAE network. The KL divergence and total loss terms are alike due to the $\beta$ coefficient which boosts the $\mathcal{L}_{kl}$ effect on the total loss. One intriguing observation involves the construction loss that begins to drop when the KL divergence has almost converged. The fluctuation of KL loss is handled conveniently in $\beta$MVUNet curves, Figure \ref{train2}, due to its structure, the number of parameters, and model size. Thereafter, the total loss of $\beta$MVUnet also converged smoothly. The $\beta$MVAE was trained during $24$ days, while it has been reduced into $12$ days for $\beta$MVUnet. 

The learning procedure is also observable from the training curves. The skipping connections, higher parameters and the net architecture of $\beta$MVUnet handled the KL divergence fluctuation. As indicated in Figure \ref{train2}, the last curve is KL divergence which has picks at the beginning of the training, and the loss is flattening very smoothly after that. However,  the KL divergence is rising and falling during the training phase in the rightmost curve of Figure \ref{train1}. It is hardly flatted at the end of the training. In Figure \ref{train2}, the KL divergence and construction curves are learning almost simultaneously, while the construction loss never is flatted as the $\beta$MVUNet curve. The construction loss is resonated with the KL divergence. When the KL divergence is flattened almost near the end, the construction loss is decreasing to some extent. This trade-off between the KL divergence and construction loss is important because the learning of each loss depends on the other. Thus, handling the fluctuation of KL divergence not only benefits the networks in posterior estimation, it correspondingly impact the object segmentation performance. 

\end{document}